%% file: acl_latex.tex
\pdfoutput=1

\documentclass[11pt]{article}

\usepackage[]{acl}
\usepackage{graphicx}
\usepackage{subcaption} 

\usepackage{fancyvrb}
\usepackage{verbatimbox}
\usepackage{booktabs} 

\usepackage{times}
\usepackage{latexsym}
\usepackage{subcaption}
\usepackage{enumerate}
\usepackage{xcolor,colortbl}
\newcolumntype{G}{>{\columncolor[gray]{0.95}}c}

\usepackage{arydshln}
\usepackage{tabularx}
\usepackage{balance}
\usepackage{tcolorbox}
\tcbuselibrary{listings, skins} 

\tcbset{
    enhanced,
    colback=white,         
    colframe=blue!75!black,
    fonttitle=\bfseries,   
    coltitle=black,        
    colbacktitle=white,    
    boxsep=3pt,            
    left=3pt,
    right=5pt,
    top=5pt,
    bottom=5pt
}

\usepackage{pgfplots}
\usepackage{tikz}
\definecolor{bblue}{HTML}{4F81BD}
\definecolor{rred}{HTML}{FFB303}
\definecolor{ggreen}{HTML}{9BBB59}
\definecolor{igreen}{HTML}{579c35}
\definecolor{ppurple}{HTML}{9F4C7C}

\pgfplotsset{
   compat=1.14,
   legend entry/.initial=,
   every axis plot post/.code={%
       \pgfkeysgetvalue{/pgfplots/legend entry}\tempValue
       \ifx\tempValue\empty
           \pgfkeysalso{/pgfplots/forget plot}%
       \else
           \expandafter\addlegendentry\expandafter{\tempValue}%
       \fi
   },
}

\usepackage[T1]{fontenc}
\usepackage{tabularray}
\usepackage[inline]{enumitem}
\usepackage[utf8]{inputenc}

\usepackage{microtype}

\usepackage{inconsolata}
\usepackage{mdframed}
\usepackage{placeins}
\usepackage{multicol}
\usepackage{keystroke}
\usepackage{menukeys}

\usepackage{multirow}
\usepackage{array}

\usepackage{cleveref}
\usepackage[symbol]{footmisc}

\newcommand{\mcqa}{\textsc{mcqa}}
\newcommand{\aqa}{\textsc{aqa}}
\newcommand{\llm}{\textsc{llm}}
\newcommand{\qalm}{\textsc{m-qalm}}
\newcommand{\liveqa}{\textsc{LiveQA}}
\newcommand{\usmle}{\textsc{USMLE}}

\newcommand{\medmcqa}{\textsc{MedMCQA}}
\newcommand{\processbank}{\textsc{ProcessBank}}
\newcommand{\pubmedqa}{\textsc{PubMedQA}}
\newcommand{\mmlu}{\textsc{MMLU}}
\newcommand{\headqa}{\textsc{HeadQA}}
\newcommand{\qamre}{\textsc{QA4MRE}}
\newcommand{\biomrc}{\textsc{BioMRC}}
\newcommand{\bioasqmcq}{\textsc{BioASQ-MCQ}}
\newcommand{\bioasqqa}{\textsc{BioASQ-QA}}
\newcommand{\mashqa}{\textsc{MashQA}}
\newcommand{\medquad}{\textsc{MedQuAD}}
\newcommand{\mediqa}{\textsc{MediQA-Ans}}
\newcommand{\medinfo}{\textsc{MedInfo}}
\newcommand{\britishopthamology}{\textsc{Ophth}}

\newcommand{\chatdoctor}{\texttt{ChatDoctor}}
\newcommand{\medalpaca}{\texttt{MedAlpaca}}
\newcommand{\pmcllama}{\texttt{PMC-LLama}}

\newcommand{\mpt}{\texttt{MPT}}
\newcommand{\falcon}{\texttt{Falcon}}
\newcommand{\lla}{\texttt{LLaMA}~1}

\newcommand{\llama}{\texttt{LLaMA}~2}
\newcommand{\flan}{\texttt{Flan-T5}}

\newcommand{\rouge}{\texttt{ROUGE-L}}
\newcommand{\bertscore}{\texttt{BERTScore}}
\newcommand{\meteor}{\texttt{METEOR}}

\newcommand{\improvement}[1]{\textcolor{igreen}{\ #1}}

\newcommand{\decrease}[1]{\textcolor{red}{\ #1}}

\title{M-QALM: A Benchmark to Assess Clinical Reading Comprehension and Knowledge Recall in Large Language Models via Question Answering}

\author{\textbf{Anand Subramanian$^{\beta}$\footnotemark}\textnormal{,} \textbf{Viktor Schlegel$^{\alpha,\gamma}$}\textnormal{,} \textbf{Abhinav Ramesh Kashyap$^{\alpha}$}\textnormal{,}\\
\textbf{Thanh-Tung Nguyen$^{\alpha}$}\textnormal{,} \textbf{Vijay Prakash Dwivedi$^{\alpha}$} \textnormal{and} \textbf{Stefan Winkler$^{\alpha,\beta}$} \\
$^{\alpha}$: ASUS Intelligent Cloud Services (AICS), Singapore\\
$^{\beta}$: National University of Singapore, Singapore\\
$^{\gamma}$: University of Manchester, United Kingdom \\
\fontsize{10}{10}\texttt{anands@u.nus.edu}, \texttt{winkler@nus.edu.sg} \\
\fontsize{10}{10}\texttt{\{viktor\_schlegel,abhinav\_kashyap,thomas\_nguyen,vijay\_dwivedi\}@asus.com}\\
}
\begin{document}
\maketitle
\begin{abstract}
There is vivid research on adapting Large Language Models (LLMs) to perform a variety of tasks in high-stakes domains such as healthcare. Despite their popularity, there is a lack of understanding of the extent and contributing factors that allow LLMs to recall relevant knowledge and combine it with presented information in the clinical and biomedical domain---a fundamental pre-requisite for success on down-stream tasks.
Addressing this gap, we use Multiple Choice and Abstractive Question Answering to conduct a large-scale empirical study on 22 datasets in three generalist and three specialist biomedical sub-domains. 
Our multifaceted analysis of the performance of 15 LLMs, further broken down by sub-domain, source of knowledge and model architecture, uncovers success factors such as instruction tuning that lead to improved recall and comprehension. We further show that while recently proposed domain-adapted models may lack adequate knowledge, directly fine-tuning on our collected medical knowledge datasets shows encouraging results, even generalising to unseen specialist sub-domains. We complement the quantitative results with a skill-oriented manual error analysis, which reveals a significant gap between the models' capabilities to simply recall necessary knowledge and to integrate it with the presented context.
To foster research and collaboration in this field we share M-QALM---our resources, standardised methodology, and evaluation results---with the research community to facilitate further advancements in clinical knowledge representation learning within language models.

\end{abstract}

\input{section/Introduction.tex}

\input{section/related_work}
\input{section/datasets_alternate}

\input{section/experiments}

\input{section/conclusion}

\section*{Limitations}
In this paper, we evaluate the medical or clinical knowledge of \llm{s} by measuring their capability of answering test questions. While this can be a useful proxy measure of a model's domain knowledge, it is insufficient to gauge its potential application in a real-world scenario. A multi-dimensional analysis of a model's behaviour, including judging the completeness, harmlessness and usefulness of generated answers, is required in addition to solely evaluating their correctness. 

Furthermore, the aggregated resource presented in this paper might be seen as lacking diversity, as all collected datasets are in English. To make inferences about the capabilities of evaluated models in other languages, a more diverse dataset with examples in other languages is required.

For our finetuning experiments, we only use parameter-efficient finetuning methods (PEFT) with QLora due to the high compute requirements for full-finetuning. We have not investigated the impact of the full-finetuning of these LLMs on our benchmark.
\bibliography{custom,references}

\clearpage

\input{section/appendix}

\end{document}

%% file: section/introduction.tex
\section{Introduction}
\renewcommand{\arraystretch}{1.3}
\begin{table*}[t!]
\centering
\footnotesize
\resizebox{\textwidth}{!}{%
\begin{tabular}{p{10cm} c  c c}

\hline
\textbf{Dataset}        & \textbf{Type}  & \textbf{Size}    & \textbf{Domain}           \\ \hline
\usmle{} \cite{medqa}                  & \mcqa{}           & 10178/1272/1273  & General Medical         \\
\medmcqa{} \cite{medmcqa}                & \mcqa{}          & 182822/4183/6150 & General Medical       \\
\bioasqmcq{} \cite{bioasq1, bioasq2}            & \mcqa{}            & 975/173/123      & General Biomedical            \\
\headqa{} \cite{headqa}                 & \mcqa{}           & 2657/1366/2742   & General Medical        \\
\processbank{} \cite{processbank}            & Context + \mcqa{} & 358/77/150       & Biological Processes \\
\pubmedqa{} \cite{pubmedqa}               & Context + \mcqa{} & 400/100/500      & General Biomedical            \\
\mmlu{} \cite{mmlu}                    & \mcqa{}           & 30/NA/1089       & General Medical/Clinical  \\
\biomrc{}-Tiny A \cite{biomrc}        & Context + \mcqa{} & NA/NA/30         & General Biomedical    \\
\biomrc{}-Tiny B \cite{biomrc}        & Context + \mcqa{} & NA/NA/30         & General Biomedical     \\
\britishopthamology{} \cite{opth, opth1, opth2}     & \mcqa{}           & NA/NA/92         & Ophthalmology         \\
\qamre{}-(Alzheimer's QA) \cite{qa4mre} & \mcqa{}           & NA/NA/40         & Alzheimer's Disease \\
\hdashline
Total Questions across Splits & - & 197420/7171/12219 & - \\
\hline
\liveqa{} \cite{liveqa, medquad}        & \aqa{}            & NA/NA/131        & Consumer Health       \\
\mediqa{} \cite{mediqa-ans}             & \aqa{}            & NA/NA/156        & Consumer Health         \\
\bioasqqa{}  \cite{bioasq1, bioasq2}             & \aqa{}            & 4733/697/363     & General Biomedical             \\
\mashqa{} \cite{mashqa}                  & \aqa{}            & 27728/3587/3493  & Consumer Health    \\
\medquad{} \cite{medquad}                & \aqa{}            & 14068/981/1358   & General Medical     \\
\medinfo{} \cite{medinfo}                & \aqa{}            & NA/NA/663        & Consumer Medication     \\
\hdashline
Total Questions across Splits & - & 46529/5265/6164 & - \\
\hline
\end{tabular}
}
\caption{Overview of \qalm{} datasets. Size is presented in terms of train/val/test splits. Manual train/val splits are created for \bioasqmcq{}, \processbank{}, \pubmedqa{}, \bioasqqa{} and \medquad. We use 6 subsets of the MMLU dataset that pertain to testing clinical and medical knowledge \cite{medpalm1}.}
\label{tab:datasets}
\end{table*}

\begin{figure}[t]
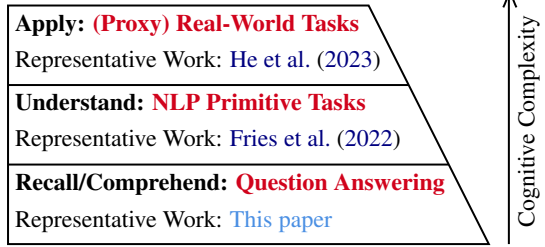

    \centering
    \include{section/figures/abstract}
    \vspace{-2em}
    \caption{The landscape of \llm{} evaluation in the medical domain with \textcolor[rgb]{0.82,0.01,0.11}{representative evaluation tasks}, organised by Bloom's taxonomy of learning objectives (bold) \cite{bloom1956taxonomy}.}
    \label{fig:SAM}
\end{figure}
\footnotetext[1]{Work done during an internship at AICS}
\renewcommand{\thefootnote}{\arabic{footnote}}
Recent success in the application of proprietary large language models in the knowledge-intensive medical domain \cite{medpalm1,medpalm2} has sparked vivid research interest in applying smaller, more readily available open-source \llm{s} to various settings in the clinical and biomedical domains. Examples of  tasks include summarization of clinical text \cite{vanveen2023clinical}, automatic note generation for physicians \cite{ben-abacha-etal-2023-empirical}, and condensation of doctor-patient dialogues \cite{mediqa2023,clinicalcamel}. More broadly, 
open-source \llm{s} have been adapted to the domain to serve as foundational clinical models \citep{medalpaca,pmcllama,clinicalcamel,biomedlm,chatdoctor}.

The success of such adaptation is typically established by measuring the performance on  down-stream tasks, by means of token overlap or semantic similarity-based metrics \cite{rouge,bertscore}. To address their inherent weaknesses \cite{Schlegel2020a,Gatt2018}, research attempts to incorporate specific dimensions, such as factuality or faithfulness \cite{medhalt}. Two important problems remain, however. Firstly, Natural Language Generation (NLG) evaluation metrics are merely approximations of the phenomena they aim to measure, and their effectiveness is typically established by the degree of correlation to human judgements of the evaluated criteria \cite{huang2021factual}. Secondly, an (offline) evaluation setup is \emph{functionally grounded} and serves as a proxy of a real-world application scenario, but the transferability of insights from functionally-grounded to application-grounded evaluation is barely discussed \cite{Doshi-Velez2017TowardsLearning}.
Taken together, these problems might taint the credibility of conclusions about the successful adaptation of \llm{s} drawn from such experiments. 

Given such difficulties, we approach the problem of evaluating \llm{} adaptation from a complementary angle. Specifically, we ask: \emph{Do \llm{s} possess the necessary pre-requisites to succeed in the clinical and medical domains}? Without an established theory of how knowledge is acquired and organised in \llm{s}, the present work is guided by the established theories of knowledge acquisition in \emph{humans}~\cite{adams2015bloom}. Typical NLG tasks, such as summarisation, are higher-level cognitives that require the understanding of learned knowledge and its application in new contexts \cite{bloom1956taxonomy}. They build on the most fundamental capability of \emph{reading comprehension}~\cite{Kintsch1988}: the construction of a text-base and its integration with previously acquired background knowledge. In NLP research, this process is evaluated by \emph{open-book Question Answering} (QA), the task of either generating (abstractive, \aqa{}) or selecting among presented options (multiple-choice, \mcqa{}) the correct answer for a question, where potentially not all necessary information is included in the question or the presented context. \mcqa{} evaluation does not suffer from the issues pertaining to NLG metrics, as performance is established by exact match. Thus, conclusions obtained from such evaluations tend to be more robust, if the quality of the benchmark is sufficient.

Therefore, in this paper we focus on the task of QA, to evaluate knowledge recall and comprehension pre-requisites of \llm{s} for successful adaptation to the medical domain.
We present an exhaustive, publicly available QA benchmark called \qalm{} including 16 \mcqa{} datasets. To enable future research on NLG-based QA, we complement \qalm{} by 6 high-quality \aqa{} datasets, where the ground-truth answer is an unconstrained string.
With such a standardized benchmark, we conduct an extensive evaluation of the capabilities of openly available general-purpose and medical \llm{s}, both ``out-of-the-box'' and after fine-tuning on \qalm{}. Our findings provide insights into the strengths and weaknesses of different \llm{s} across a range of datasets, question categories and QA tasks. Overall, we find their performance lacking, both compared to humans and to proprietary \llm{s}. Further analysis reveals promising tendencies of domain-specific pre-training and fine-tuning to bridge this gap and to generalise to new QA datasets.

%% file: section/figures/abstract.tex
\tikzset{every picture/.style={line width=0.75pt}}        

\begin{tikzpicture}[x=0.75pt,y=0.75pt,yscale=-1,xscale=1]
\draw    (10,110) -- (230,110) ;
\draw    (10,70) -- (210,70) ;
\draw    (260,150) -- (260,22) ;
\draw [shift={(260.5,20)}, rotate = 90] [color={rgb, 255:red, 0; green, 0; blue, 0 }  ][line width=0.75]    (10.93,-3.29) .. controls (6.95,-1.4) and (3.31,-0.3) .. (0,0) .. controls (3.31,0.3) and (6.95,1.4) .. (10.93,3.29)   ;
\draw   (190,30) -- (250,150) -- (10,150) -- (10,30) -- cycle ;

\draw (12,113) node [anchor=north west][inner sep=0.75pt]  [font=\footnotesize] [align=left] {\bfseries Recall/Comprehend: \textcolor[rgb]{0.82,0.01,0.11}{Question Answering}};
\draw (12,73) node [anchor=north west][inner sep=0.75pt]  [font=\footnotesize] [align=left] {\bfseries Understand: \textcolor[rgb]{0.82,0.01,0.11}{NLP Primitive Tasks}};
\draw (12,33) node [anchor=north west][inner sep=0.75pt]  [font=\footnotesize] [align=left] {\bfseries Apply: \textcolor[rgb]{0.82,0.01,0.11}{(Proxy) Real-World Tasks}};
\draw (47,160) node  [font=\footnotesize] [align=left] {\textit{"Know-What"}};
\draw (45,18) node  [font=\footnotesize] [align=left] {\textit{"Know-How"}};
\draw (270,88) node  [font=\footnotesize,rotate=-270] [align=left] {Cognitive Complexity};
\draw (12,132) node [anchor=north west][inner sep=0.75pt]  [font=\footnotesize] [align=left] {Representative Work: \textcolor[rgb]{0.29,0.56,0.89}{This paper}};
\draw (12,98) node [anchor=west] [inner sep=0.75pt]  [font=\footnotesize] [align=left] {Representative Work: \citet{fries2022bigbio}};
\draw (12,58) node [anchor=west] [inner sep=0.75pt]  [font=\footnotesize] [align=left] {Representative Work: \citet{he2023medeval}};

\end{tikzpicture}

%% file: section/related_work.tex
\section{Related Work}

\textbf{Large open-domain QA benchmarks} The availability of QA datasets from multiple domains and sources has enabled the curation of large and diverse QA benchmarks \cite{Dua2019a,Fisch2019MRQAComprehension,Talmor2019}. Such resource collections enable researchers to perform large-scale empirical studies to understand how well language models can generalise to new questions from new domains or sources, or how fine-tuning can impact this performance. While multiple studies exist in the general domain, to the best of our knowledge, no such large-scale study has been carried out for QA in the clinical domain. In this paper we aim to address this gap.

\textbf{Evaluation in the clinical domain} 
Datasets that evaluate the lowest-level cognitive task of knowledge recall and reading comprehension in the medical domain have been proposed before \cite{medqa,headqa,medmcqa}. They feature questions commonly found in examinations like the US Medical Licensing Exam (USMLE). \qalm{} unifies the existing literature by incorporating licensing exam questions from diverse regions, such as India and Spain. We go beyond the scope of the general medical domain, covering specialist topics such as ophthalmology and Alzheimer's disease.

Beyond factual recall and comprehension, \citet{fries2022bigbio} collected a unified bio-medical benchmark, featuring NLP primitives such as \mbox{sentence(-pair)} classification or entity recognition and linking. Aiming at higher, more task-specific cognitives, \citet{medpalm1} introduced MultiMedQA, including HealthSearchQA, which requires models to generate high-quality free-form answers. Similarly, \citet{he2023medeval} introduced a multi-domain benchmark for evaluating generation and classification capabilities on a diverse set of in-hospital downstream tasks. Other researchers looked to evaluate the quality and factuality of generation \cite{medhalt} and synthesised general-purpose medical instructions \cite{fleming2023medalign}.
Our work is complementary, because we evaluate knowledge recall and comprehension as a pre-requisite of higher-level cognitive tasks, such as understanding and application---the focus of previously discussed works.

%% file: section/datasets_alternate.tex
\section{\qalm{} Datasets}

The primary goal of \qalm{} is to develop a comprehensive, open-source repository of medical QA datasets to assess the recall of medical knowledge in \llm{s}. To obtain such a collection, we perform an exhaustive literature and resource search using the terms ``clinical OR medical'', ``Question Answering OR QA'' and include a dataset or resource if it satisfies the following criteria:
\begin{enumerate*}[label=(\itshape \roman*)]
\item The language is English, as medical documents are usually written in English, even in non-English-speaking countries;
\item The questions and answers are on general, specialist, or consumer-facing medical topics;
\item The resource is openly available without restrictive licensing or data agreements;
\item The resource evaluates the task of \mcqa{} or \aqa{};
\item The ground truth is collected or reviewed by domain experts.
\end{enumerate*}

The result is  \qalm{}---a comprehensive collection of 22 datasets designed to thoroughly evaluate the clinical knowledge of \llm{s}. \Cref{tab:datasets} gives an overview of the collected \mcqa{} and \aqa{} datasets, including task formulation, size and domain. Refer to the Appendix for further details on each dataset.

\textbf{Knowledge source categorization} The \mcqa{} datasets within the \qalm{} benchmark cover a diverse range of medical domains. 
To be able to perform fine-grained analysis of both the topics covered in these datasets as well as  model performance, we categorise the \mcqa{} datasets into eleven high-level categories, representing different facets of medical knowledge. 
To do so, we leverage available meta-data from the source datasets \medmcqa{}, \headqa{}, \mmlu{} and \bioasqmcq{}. We categorize the \processbank{}, \pubmedqa{} and \biomrc{} datasets into a distinct twelfth \texttt{Within Context} category, as the relevant knowledge is presented in the context.  \usmle{} and \qamre{} lack the necessary meta-data, thus we train a BioBERT-based classifier \cite{biobert} to assign questions into one of the eleven elicited categories using the labels from the other datasets. The classifier achieves 71.56\% \mbox{(micro-)averaged} F1 score on a held-out test set, which we deem sufficient.

Table~\ref{tab:results-categories} shows that nearly half of all questions (47\%) fall into the \texttt{Basic and Life Sciences} and \texttt{General Medicine} category. \texttt{Diagnostic Sciences}, \texttt{Women's and Children's Health} and \texttt{Pharmacology and Anesthesia} account for a further 30\% of questions.

%% file: section/experiments.tex
\section{Empirical Evaluation}
We investigate how well existing, open-source \llm{s} are able to recall clinical knowledge and integrate it into a given context in order to succeed on our benchmark. Specifically, we focus on performance in the zero-shot setting, and after fine-tuning on \qalm{} training portions. \\
In the \textbf{Zero-shot} setting: 
    \begin{itemize}[leftmargin=10pt,noitemsep,topsep=0pt]
    \item \textbf{RQ1.} How well do open-source \llm{s} recall necessary clinical knowledge when they are tested on \qalm{}?
    \item \textbf{RQ2.} Does open-domain instruction fine-tuning of \llm{s} improve their ability to do so?
    \item \textbf{RQ3.} Does \emph{domain-specific} fine-tuning improve performance on \qalm{}?
    \end{itemize}
In the \textbf{Fine-tuned} setting:
    \begin{itemize}[leftmargin=10pt,noitemsep,topsep=0pt]
    \item \textbf{RQ4.} Does finetuning on \qalm{} improve performance on unseen data from datasets seen during training?
    \item \textbf{RQ5.} Does fine-tuning improve performance on \emph{unseen} \qalm{} datasets?
    \end{itemize}

\subsection{Study Setup}
To seek evidence for \textbf{RQs 1-3} empirically, we evaluate several \llm{s} and their instruction-tuned versions on the test splits of \qalm{} in zero-shot\footnote{For \mcqa{} evaluation in the zero-shot setting (where models are not explicitly fine-tuned for \mcqa{} tasks), we use a 1-shot prompt---giving an example to the model, and find that it adheres better to the \mcqa{} format and the standard 5-shot prompt for MMLU datasets.} manner. To answer \textbf{RQ4} and \textbf{RQ5}, we fine-tune \llm{s} on the training portion of \qalm{} and evaluate on test splits of datasets both seen and unseen during training. We complement our evaluation with additional automated and manual error analyses to identify causes for model successes and failures.

\textbf{Models:} To assess the zero-shot capabilities of models (\textbf{RQ1} and \textbf{RQ2}), we include a diverse array of open-source decoder-only models with parameter scales ranging from 3B-13B. We use models from \mpt{} and \mpt{}-Instruct (7B) \cite{mpt}, \falcon{} and \falcon{}-Instruct (7B) \cite{falcon}, \lla{} (7B and 13B) \cite{llama}, \llama{} and \llama{}-chat (7B and 13B) \cite{llama2}.
In addition to these models, we also use two instruction fine-tuned encoder-decoder models: \flan{} (3B and 11B) \cite{flant5}. Models with \textit{Instruct} or \textit{Chat} appended to their names are instruction fine-tuned \cite{instructgpt} versions of their base models. The details of the models are given in Table \ref{tab:models}. To address \textbf{RQ3}, we evaluate \chatdoctor{} (7B) \cite{chatdoctor}, \medalpaca{} (7B) \cite{medalpaca}  and \pmcllama{} \cite{pmcllama}. To address \textbf{RQ4}, we fine-tune models using the training set of the \qalm{} datasets. When official validation splits are unavailable, we employ a random split of up to around 20\% of the data for validation purposes. If no training datasets are available, we do not use this dataset for fine-tuning and only consider the test split of the respective datasets to answer \textbf{RQ5}. For evaluating \aqa{}, we use a sub-sampled version of the test sets of \mashqa{} (500 questions) and \medquad{} (200 questions by sampling 100 questions from the two holdout websites), while we use the other datasets as they are. For \mcqa{}, similar to \citet{medpalm1}, we evaluate all models on the validation set of \medmcqa{} since the answers for the test set are not released publicly.

\textbf{Finetuning and hyperparameters:} Since the number of parameters for most of our models is in the billions, we follow a more accepted practice of using parameter-efficient fine-tuning, specifically  QLora and 4-bit quantization \cite{qlora}. We utilize 8-bit quantization for evaluating \flan{} (11B), \lla{} (13B), \llama{} (13B) and \llama{}-Chat (13B) \cite{int8}. We use A100-40G GPUs for all our experiments. The other hyper-parameters used to train our models are reported in the Appendix (\Cref{tab:hyperparams}).

\textbf{Evaluation measures:} We use Accuracy to measure the performance of the model on \mcqa{} datasets; for \aqa{} datasets, we use \rouge{}  \cite{rouge}, \bertscore{}  \cite{bertscore} (based on deberta-xlarge-mnli) and \meteor{} \cite{meteor}, which is found to correlate better with human judgments than other metrics on \aqa{} \cite{meteor_correlation}.

\section{Results and Analysis}
In this section, we report and analyse the findings of our empirical study.

\begin{table}[!t]
    \centering
    \large
    \resizebox{.99\columnwidth}{!}{%
    
    \renewcommand{\arraystretch}{1.25}
    \begin{tabular}{p{0.03cm} l | c : G c G c}
    &  & \multicolumn{1}{c}{\textbf{{\mcqa{}}}} &  \multicolumn{3}{c}{\textbf{{\aqa{}}}} & \\
    &  & Acc & RL & BS & MTR \\
        \hline
        \multirow{6}{*}{\rotatebox[origin=c]{90}{\emph{Base}}} &
 \lla{} (7B)          & 31.9 & 14.0 & 54.2 & 20.5 & \\
& \lla{} (13B)          & 44.1 & 14.4 & 54.0 & 20.3 & \\ &
 \llama{} (7B)          & 42.9 & 14.9 & 55.3 & 21.1 & \\
& \llama{} (13B)          & 47.1 & 15.0 & 56.4 & 22.5 & \\

& \mpt{} (7B)          & 27.6 &13.3 &52.6 & 21.1 & \\
& \falcon{} (7B)          & 34.7 & 14.0 & 54.1 & 20.0 & \\
\hdashline
\multirow{6}{*}{\rotatebox[origin=c]{90}{\emph{Instruction tuned}}}
& \llama{}-chat (7B)       & 45.9 & 15.0 & 58.0 & 23.3 & \\
& \llama{}-chat (13B)          & 50.3 & 15.3 & 58.0 & 23.6 & \\
& \mpt{}-Instruct (7B)          & 31.6 & 15.8 & 59.7  & 15.6 & \\
& \falcon{}-Instruct (7B)          & 31.8 & 17.2 & 62.4 & 17.4 & \\
& \flan{} (3B)          & 51.8 & 10.8 & 55.0 & 7.4 & \\
& \flan{} (11B)          & 56.5 & 11.5 & 56.3 & 8.2 & \\
\hdashline
\multirow{3}{*}{\rotatebox[origin=c]{90}{\emph{Adapted}}}
& \chatdoctor{} (7B)      & $42.8$ & $17.4$ & $62.3$ & $18.7$ &   \\
& \medalpaca{} (7B)      & $48.8$ & $15.5$ & $58.9$ & $15.6$  \\
& \pmcllama{} (13B)      & $53.7$ & $19.7$ & $60.7$ & $19.0$  \\
\hdashline
\multirow{1}{*}{\rotatebox[origin=c]{90}}
& Random Baseline      & $27.7$ & - & - & - &   \\
\hline
     \end{tabular}
     }
    \caption{Zero-shot performance of base (top), instruction-tuned models (middle) and domain-adapted (bottom) models. Metrics are \textbf{Acc}uracy for \mcqa{}; \textbf{R}ouge-\textbf{L}, \textbf{B}ERT\textbf{S}core, and \textbf{M}E\textbf{T}EO\textbf{R} for \aqa{}.}
    \label{tab:results}
\end{table}

\subsection{Zero-shot Evaluation Results}
\Cref{tab:results} shows the dataset-averaged scores of the zero-shot evaluation of language models as evidence towards \textbf{RQs 1-3}. Note that in this way, each dataset contributes equally to the average, regardless of its size. Micro-averaged \mcqa{} accuracy scores are reported in Table~\ref{tab:results-categories}. However, these are biased towards datasets with more examples (i.e., \medmcqa{}). While the results between micro- and by-dataset-averaged metrics might differ in detail (consult Appendix~\ref{sec:appendix-all-details} for a break-down), the mean absolute difference between the metrics for all models is 4.2, which suggests that reported trends do not depend on the averaging method.

\paragraph{}
\Cref{tab:results} highlights that \llm{s} exhibit \textbf{strong zero-shot capability on \mcqa{} and \aqa{} datasets}, corroborating  the findings of \citet{medpalm1}. Considering \llm{s} of the same size (e.g., 7B), \llama{} performs best, 
possibly due to larger diversity in pre-training data---\llama{} is trained on the most tokens. Another difference is the mixture of datasets used for pre-training, which is not revealed in some cases (c.f. \Cref{tab:models} in Appendix).

Unsurprisingly, across all models of the same architecture, \textbf{scale predicts model performance}, even without domain-specific adaptation of \llm{s} on the medical domain. 
For example, \llama{} (13B) performs better on \mcqa{} ($+4.2$ Accuracy improvement) compared to the 7B version. 
Figure~\ref{fig:by-params} in the Appendix shows the relationship between the number of parameters and performance.

\paragraph{} 
To address \textbf{RQ2}, we investigate whether improvements from instruction fine-tuning also apply to the clinical domain of \qalm{}. The results are reported in the middle part of \Cref{tab:results}.

Surprisingly, \textbf{instruction fine-tuned models perform better} than their corresponding \emph{Base} versions, despite the fact that the instruction set used for fine-tuning contains only tasks in the general domain---see \Cref{tab:models} and compare *-Instruct/Chat (middle) with their base versions (top). Among them, \flan{} models exhibit the best zero-shot performance on \mcqa{}, outperforming comparable decoder-only models. Seemingly, instruction fine-tuning enables models to obtain  representations of question and context which are beneficial for fact recall.

We note that \textbf{bigger models are not always better}---the choice of model architecture and  dataset for instruction fine-tuning can have a bigger impact on performance than model size alone. For example the encoder-decoder \flan{} (3B) model outperforms \llama-chat (13B) on the \mcqa{} task, despite being four times smaller.

\paragraph{}
The performance of domain-adapted models is reported in \Cref{tab:results} (bottom), as evidence for \textbf{RQ3}.
For \mcqa{}, both \medalpaca{} and \chatdoctor{} indeed exhibit improvements in Accuracy over their respective 7B and 13B \lla{} base versions; however they fail to reach the strong zero-shot performance of \flan{} (11B).

In contrast, \pmcllama{} performs well due to continued pre-training on biomedical corpora before instruction tuning on biomedical and clinical datasets. The latter results in exceptionally high scores on the \medinfo{} \aqa{} dataset (See  \Cref{tab:aqa_scores} in Appendix). This dataset, along with \liveqa{}, was used as part of the instruction tuning process, leading to evaluation on these dataset not being ``zero-shot''\footnote{\url{https://huggingface.co/datasets/axiong/pmc\_llama\_instructions}}. Scores on \liveqa{}, however, are not inflated, compared to \llama{}(-chat) (13B). This is possibly because  we use a filtered version of \liveqa{} which contains only challenging answers with sufficiently good expert quality rating. \pmcllama{} demonstrates significant improvements over other open-source LLMs on \mcqa{} datasets such as \usmle{}, \medmcqa{} and \mmlu{}.

In summary, we conclude that while available \llm{s} adapted to the medical domain successfully improve performance of the adapted models, they appear to have \textbf{no improved domain knowledge} compared to other available open-domain models. Evaluating these adaptation techniques on stronger base models is an exciting avenue for future research.

Importantly, none of the evaluated
open-source \llm{s} outperform humans: While the passing score for USMLE is  60\%
\footnote{\url{https://www.usmle.org/bulletin-information/scoring-and-score-reporting}}, we observe the best zero-shot scores for USMLE are 43\% for \llama{}, and 54\% for the domain-adapted \pmcllama{}, both below the passing score. 
Meanwhile, GPT-4 \cite{gpt4} with a customized prompting strategy labeled MedPrompt \cite{medprompt} achieves 90.2\%, while Med-PALM 2 \cite{medpalm2} achieves scores of 86.5\% on USMLE. Similarly, for the PubmedQA dataset, human performance is 78\% \cite{pubmedqa}, compared to 72.4\% of \flan{}. To summarize: While available \llm{s} exhibit performance significantly higher than random chance ``out-of-the-box'', there is still a \textbf{substantial gap compared to humans and proprietary \llm{s}} \citep{medpalm1,medpalm2} (see \Cref{sec:appendix-best-perf}).

\subsection{Impact of Fine-tuning}

Given the scale of \qalm{}, we are able to fine-tune models on parts of the data, to address \textbf{RQ4} and \textbf{RQ5}. We fine-tune four models on \mcqa{} and \aqa{} separately, given the different nature of these datasets, but joint fine-tuning on both \mcqa{} and \aqa{} did not yield significantly different results.

\begin{table}[hb!]
    \centering
    \resizebox{\columnwidth}{!}{%
    \renewcommand{\arraystretch}{1.5}
    \begin{tabular}{l | c : G c G c}
      & \multicolumn{1}{c}{\textbf{{\mcqa{}}}} &  \multicolumn{3}{c}{\textbf{{\aqa{}}}} \\
    & Acc & RL & BS & MTR \\
        \hline

\llama{} (7B)         & $53.5_{\improvement{+10.6}}$ & $17.7_{\improvement{+2.8}}$ & $60.8_{\improvement{+5.5}}$ & $16.9_{\decrease{-4.2}}$ \\
\falcon{} (7B)       & $49.3_{\improvement{+14.6}}$ & $17.4_{\improvement{+3.4}}$ & $60.4_{\improvement{+6.3}}$ & $17.1_{\decrease{-2.9}}$ \\
\mpt{} (7B)          & $53.2_{\improvement{+25.6}}$ & $17.3_{\improvement{+4.0}}$ & $60.0_{\improvement{+7.4}}$ & $17.2_{\decrease{-3.9}}$ \\
\flan{} (3B)      & $52.9_{\improvement{+1.1}}$ & $15.9_{\improvement{+5.1}}$ & $56.8_{\improvement{+1.8}}$ &  $15.6_{\improvement{+8.2}}$ \\
\hline
     \end{tabular}
     }
    \caption{Model fine-tuning is performed either on \mcqa{} or \aqa{} datasets. Reported are \textbf{Acc}uracy for \mcqa{}; \textbf{R}ouge-\textbf{L}, \textbf{B}ERT\textbf{S}core, and  \textsc{\textbf{M}E\textbf{T}EO\textbf{R}} for \aqa{}. Subscripts indicate improvement over zero-shot versions.}
    \label{tab:results-ft}
\end{table}

We fine-tune the models only on the \mcqa{} subset of datasets first (cf. \Cref{tab:results-ft}). We find that the \textbf{fine-tuned models perform better} compared to their non-fine-tuned counterparts. Decoder-only models like \mpt{} (7B) benefit more than others ($+25.6$ Accuracy improvement).  
Fine-tuning models on the data seems to close the gaps introduced by different model architectures and pre-training data: 
The standard deviation of the evaluated models' accuracies reduces from $9.0$ in the zero-shot setting to $1.7$ after fine-tuning.  This suggests that \llm{s} can benefit from task-specific fine-tuning to address seemingly sub-optimal architecture or pre-training conditions. For \aqa{}, \flan{} benefits more from fine-tuning compared to the decoder-only models, possibly by better aligning generated outputs to the expected format of the answer.
Decoder models present inconsistent results with improvements in \rouge{} and \bertscore{} at the expense of lower \meteor{} scores, which raises concerns about the reliability of the \aqa{} metrics.

\begin{table*}[ht!]
    \centering
    \resizebox{\textwidth}{!}{
    \renewcommand{\arraystretch}{1}
    \begin{tabular}{l l c c c c c c c c c c}
 & \\ \hline
  &   &  & \flan{} & \flan{} & \mpt{} & \mpt{} & \falcon{} & \falcon{} & \llama{} & \llama{} \\
  &  Category & Support & (ZS) & (FT) & (ZS) & (FT) & (ZS) & (FT) & (ZS) & (FT) \\
        \hline
\multirow{3}{*}{\rotatebox[origin=c]{90}{\emph{Domain}}}
& General Medical                          &  9275        & 37.9       & 44.9            & 26.5                          & 49.5         & 29.5         &  46.9          & 37.6                              &   \textbf{50.5}        \\
& General Biomedical                          &  683        & 64.4       & \textbf{71.0}            &  32.4                         & 70.0         & 56.7         & 68.4           & 58.9                              &  68.5         \\
& Biological                          &  294        & \textbf{71.4}        & 70.4             & 39.5                           & 71.1         &  39.1        & 58.2           &  57.8                             & 68.4          \\
\hline
\multirow{14}{*}{\rotatebox[origin=c]{90}{\emph{Knowledge Source}}}
& General Medicine                   & 2675          & 38.0         & 43.2            & 26.0                         & 46.4        & 30.1        & 46.4           & 36.6                             & \textbf{50.0}           \\
& Basic and Life Sciences            & 2235          & 38.9         & 44.3            & 26.9                         & \textbf{52.6}        & 30.6        & 49.4           & 40.0                             & 52.5           \\
& Dental and Oral Health             & 1318          & 34.8         & 42.9            & 25.9                         & \textbf{44.3}        & 30.7        & 43.8           & 36.1                             & 44.2           \\
& Pharmacology and Anesthesia        & 784           & 39.7         & 48.1            & 29.0                         & 55.6        & 28.8        & 54.0           & 42.9                             & \textbf{59.4 }          \\
& Within Context              & 710           & 74.1         & \textbf{75.2}            & 37.2                         & 71.5        & 52.7        & 66.5           & 60.8                             & 67.7           \\
& Diagnostic Sciences                & 640           & 32.2         & 43.1            & 26.4                         & \textbf{51.1}        & 30.3        & 46.4           & 37.2                             & 47.5           \\
& Supportive and Preventive Services & 599           & 48.2         & \textbf{56.6 }           & 23.7                         & 55.1        & 27.9        & 48.1           & 39.9                             & 56.3           \\
& Women's and Children's Health      & 507           & 30.2         & 42.6            & 27.2                         & \textbf{51.7}        & 28.4        & 43.0           & 34.3                             & 49.9           \\
& Mental and Behavioral Health       & 496           & 50.0         & 57.9            & 29.4                         & 55.4        & 31.5        & 49.2           & 40.7                             & \textbf{59.1}           \\
& Sensory Organs                     & 205           & 29.8         & 42.0            & 27.8                         & \textbf{45.4}        & 28.8        & 42.4           & 33.2                             & 42.0           \\
& Miscellaneous                      & 45            & 42.2         & 44.4            & 20.0                         & \textbf{60.0}        & 24.4        & 44.4           & 31.1                             & 40.0           \\
& Musculoskeletal and Dermatology    & 38            & 18.4         & 26.3            & 18.4                         & \textbf{44.7}        & 34.2        & 42.1           & 28.9                             & \textbf{44.7}           \\ 
\cline{2-12}
& Micro-averaged Accuracy                          & 10252         & 40.6        & 47.4            & 27.3                         & 51.5        & 31.6        & 48.6           & 39.6                             & \textbf{52.2}          \\
& Category-averaged Accuracy                          &  12        & 39.7        & 47.2            & 26.5                          & \textbf{52.8}        &  31.5       & 48.0           & 38.5                             & 51.1          \\

\hline
     \end{tabular}
         }
\caption{Performance of \llm{s} in the zero-shot and fine-tuned setting across various categories on the test set.}

\label{tab:results-categories}

\end{table*}

Scaling up models introduces practical problems of deploying the model in real-world scenarios---smaller models may be preferred to larger ones due to faster inference times and lower memory footprints. 
We find that \textbf{fine-tuning helps compensate for scale}. Fine-tuned \llama{} (7B) significantly outperforms zero-shot \llama{} (13B) ($+6.4$ Accuracy gain on \mcqa{}, $+2.7$ \rouge{} gain and $+4.4$ \bertscore{} gain on \aqa{}). Similarly, fine-tuned \flan{} (3B) outperforms zero-shot \llama{} (13B) on 8 out of 16 \mcqa{} datasets (see Tables~\ref{tab:mcqa_zs}~and~\ref{tab:mcqa_ft}).

In summary, we conclude that task-specific fine-tuning improves performance, mitigating weaknesses due to size, architecture and training data.

\begin{figure}[t!]
\centering
    \begin{tikzpicture}
    \begin{axis}[
        ybar=0.5pt,
        bar width=3.5pt,
        x tick label style={rotate=23, font=\scriptsize\itshape},
        y tick label style={rotate=90, font=\scriptsize\itshape},
        height=10em,
        ylabel=\meteor{},
        width  = \columnwidth,
        x tick label as interval,
        ymajorgrids=true,
        y grid style=dashed,
        major y tick style = transparent,
        xtick={0, 1, 2, 3, 4, 5, 6, 7},
        xmajorgrids=true,
        xminorgrids=true,
        legend pos=north west,
        legend cell align={left},
        legend columns=2,
        xmin=0, xmax=4,
        ymin=0.0, ymax=0.5,
        xticklabels={\liveqa{}, \mediqa{}, \medquad{}, \medinfo{}}
    ]
    \addplot[legend entry=\textsc{\scriptsize LLama-ZS}, color=pink, fill=pink]  coordinates {(0.5, 0.201) (1.5, 0.223) (2.5, 0.194) (3.5, 0.178) };
    \addplot[legend entry=\textsc{\scriptsize LLAma-\aqa{}-FT},, color=teal, fill=teal]  coordinates {(0.5, 0.192) (1.5, 0.112) (2.5, 0.166) (3.5, 0.090) };
    \addplot[legend entry=\textsc{\scriptsize Flan-T5-\aqa{}-ZS},, color=bblue, fill=bblue]  coordinates {(0.6, 0.061) (1.6, 0.062) (2.6, 0.06) (3.6, 0.076) };
    \addplot[legend entry=\textsc{\scriptsize Flan-T5-\aqa{}-FT},, color=rred, fill=rred]  coordinates {(0.6, 0.169) (1.6, 0.120) (2.6, 0.127) (3.6, 0.105) };
    \end{axis}
    \end{tikzpicture}

    \caption{Performance of base and \aqa{}-fine-tuned \llama{} and \flan{} models on unseen \aqa{} test sets.}
    \label{fig:generalisation-aqa}
\end{figure}
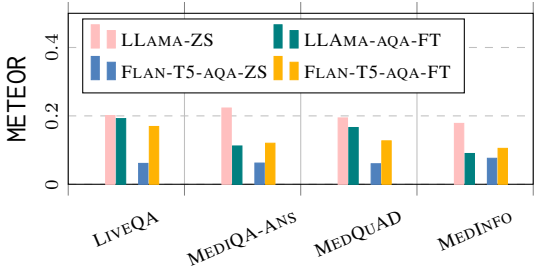

\begin{figure}[!tbh]
\centering
    \begin{tikzpicture}
    \begin{axis}[
        ybar=0.5pt,
        bar width=3.5pt,
        x tick label style={rotate=32, font=\scriptsize\itshape},
        y tick label style={rotate=90, font=\scriptsize\itshape},
        height=11.7em,
ylabel=\emph{Accuracy},
        width  = \columnwidth,
        ymajorgrids=true,
        y grid style=dashed,
        major y tick style = transparent,
        xtick={0, 1, 2, 3, 4, 5, 6, 7, 8, 9, 10},
        xmajorgrids=true,
        xminorgrids=true,
        legend pos=north west,
        legend cell align={left},
        legend columns=3,
        xmin=0, xmax=10,
        ymin=0.0, ymax=0.8,
        xticklabels={\biomrc{}-A,\biomrc{}-B, \qamre{}, \mmlu{}-AN, \mmlu{}-CK, \mmlu{}-CB, \mmlu{}-CM, \mmlu{}-MG, \mmlu{}-PM, \textsc{Ophth}}
    ]
    \addplot[legend entry=\textsc{\scriptsize ZS}, color=bblue, fill=bblue]  coordinates {(0.5, 0.3) (1.5, 0.267) (2.5, 0.4) (3.5, 0.407) (4.5, 0.381) (5.5, 0.396) (6.5, 0.353) (7.5, 0.490) (8.5, 0.441) (9.5, 0.272)};
    \addplot[legend entry=\textsc{\scriptsize \mcqa{}-FT},, color=ggreen, fill=ggreen]  coordinates {(0.5, 0.233) (1.5, 0.267) (2.5, 0.5) (3.5, 0.541) (4.5, 0.596) (5.5, 0.611) (6.5, 0.520) (7.5, 0.620) (8.5, 0.596) (9.5, 0.315)};
    \addplot[legend entry=\textsc{\scriptsize \aqa{}-FT},, color=rred, fill=rred]  coordinates {(0.5, 0.167) (1.5, 0.167) (2.5, 0.150) (3.5, 0.385) (4.5, 0.408) (5.5, 0.389) (6.5, 0.376) (7.5, 0.490) (8.5, 0.467) (9.5, 0.196)};
    \end{axis}
    \end{tikzpicture}
    
    \caption{Performance of base, \mcqa{}-tuned, and \aqa{}-tuned \llama{} model on unseen \mcqa{} test sets.}
    \label{fig:generalisation-mcq}
\end{figure}
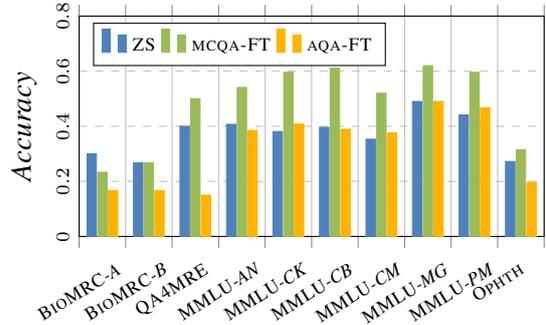

\paragraph{}
Finally, we report the potential of \llm{s} fine-tuned on in-domain data to generalize to medical datasets unseen during training to answer \textbf{RQ5}. To this end, during fine-tuning, we hold out 
ten \mcqa{} and four \aqa{}  datasets presented in Figures~\ref{fig:generalisation-aqa}~and~\ref{fig:generalisation-mcq}.

Figure~\ref{fig:generalisation-aqa} shows the performance of \llama{} (7B) and \flan{} (3B) models on the four held-out \aqa{} evaluation sets. While \llama{} does not appear to generalise to unseen \aqa{} datasets, \flan{}'s scores improve across the board. We note however, that this result might depend on the choice of metric, as Figures~\ref{fig:generalisation-aqa-rouge}~and~\ref{fig:generalisation-aqa-bs}~in the Appendix paint a more mixed picture.  Indeed, across all conducted experiments, only \rouge{} scores show a statistically significant Spearman rank correlation with the reliable \mcqa{} accuracy measure ($r=0.616$, $p=0.008$, more details in Appendix~\ref{sec:appendix-corr}). This suggests that other metrics used are either a sub-optimal choice or that they measure another, complementary aspect captured neither by Accuracy nor \rouge{}. These findings highlight the \textbf{low robustness of overlap-based NLG metrics} discussed in the introduction.

Investigating the more robust \mcqa{} setting, \Cref{fig:generalisation-mcq} (comparing blue ZS with orange AQA-FT bars) shows that \textbf{fine-tuning on \aqa{} does not improve performance on unseen \mcqa{} datasets}. This suggests that higher scores on unseen \aqa{} datasets might stem from better aligning generations to the expected answer form of \aqa{} answers, which shows improvements in some of the \aqa{} metrics, rather than acquiring additional medical knowledge during fine-tuning. While this could also due to a domain shift between the training and holdout datasets, this is not supported by the performance drop on MedQuaD, which, by this theory, should exhibit improved performance, since its domain is “General Medical”, and would thus be in-domain.

\Cref{fig:generalisation-mcq} (comparing blue ZS with green MCQ-FT) suggests that models indeed can learn to acquire domain-specific knowledge during fine-tuning, as \mcqa{}-tuned models consistently perform better than their zero-shot counterparts. This seemingly contradicts the previous finding that models fail to acquire additional medical knowledge when fine-tuned on the \aqa{} datasets.

\paragraph{}

\begin{table*}[ht!]
    \centering
    \resizebox{0.9\textwidth}{!}{
    \renewcommand{\arraystretch}{1}
    \begin{tabular}{l c c c c c c c c c c}
 & \\ \hline
    &  & \flan{} & \flan{} & \mpt{} & \mpt{} & \falcon{} & \falcon{} & \llama{} & \llama{} \\
   Reasoning Type & Support & (ZS) & (FT) & (ZS) & (FT) & (ZS) & (FT) & (ZS) & (FT) \\

        \hline
        \hline
Recall                         &  131        & 48.1       & 49.6  &          23.7        &  \textbf{51.1} & 31.3                          & 49.6                   & 47.3                              &   \textbf{51.1}        \\
Reading Comprehension                          &  59        & 27.1       & 39.0            &  27.1       & 35.6          & 40.7                       & \textbf{47.5}         &  33.9                              &  42.4       \\
Quantitative/Arithmetic                          &  10       & \textbf{40.0}        & 30.0      &       10.0 &       20.0           &  30.0                           & \textbf{40.0}         &   30.0                             & 30.0          \\

\hline
     \end{tabular}
         }
\caption{Performance of \llm{s} in zero-shot and fine-tuned settings on three reasoning types identified in \qalm{}.}
\label{tab:manual-analysis-all}
\end{table*}

Further analysis indicates that the reported generalisation capabilities might be over-stated, as evaluation questions from the unseen datasets have semantically similar counterparts in the fine-tuning data. However, a manual analysis of the cases where fine-tuned models outperform their zero-shot counter-parts reveals that only about 60\% of the improvement can be explained by the presence of such similar examples. Details of this analysis are reported in Appendix~\ref{sec:appendix-gen-analysis}.

Based on these findings, we conclude that \textbf{fine-tuning can serve as a partial solution for achieving generalisable adaptation to the medical domain}.

\section{Error Analysis}
In this section, we analyze the errors of LLMs on \mcqa{} datasets.  

\subsection{Category-wise and Manual Error Analysis}
To better understand the performance of zero-shot and fine-tuned models across \mcqa{}, we analyze them broken down by sub-domain and knowledge source. We calculate the accuracy of the models in their zero-shot and fine-tuned settings for each category, as shown in Table \ref{tab:results-categories}.

Models tend to perform better on the biological and biomedical sub-domains. We posit as the reason for this that biomedical information is more readily available in the pre-training corpora of the models, e.g., in the form of biomedical abstracts (see also Table~\ref{tab:models} in the Appendix). Furthermore, fine-tuning improves performance for all categories, but the gaps between medical and biomedical domains  persist, indicating that medical questions are indeed harder to answer, even though they prevail in the training set. Perhaps more worryingly, the Consumer Health \aqa{} scores do not improve as much as for other domains, even after fine-tuning (see Appendix, Table~\ref{tab:results-rouge-categories}).
 
For knowledge sources, fine-tuned \flan{} (3B) excels in \texttt{Within Context} and \texttt{Supportive and Preventive Services}, also showing strong zero-shot capabilities in these categories, perhaps due to architecture or pre-training data. 
Similarly, fine-tuned \mpt{}~(7B) and \llama{}~(7B) show superior performance across categories. However, despite fine-tuning benefits, models still underperform in areas like \texttt{General Medicine}, \texttt{Basic and Life Sciences}, and \texttt{Dental and Oral Health}, which form the majority of the benchmark. Overall, we conclude that \textbf{fine-tuning improves model performance in sub-domains, but knowledge gaps still persist across different domains and knowledge sources}.

Finally, we sample 200 \mcqa{}-questions from \qalm{} evaluation data, and annotate the type of reasoning required to solve the problem: we distinguish three broad categories: \texttt{Recall} questions, which only require to recall necessary knowledge, \texttt{Reading Comprehension} questions, which require recall of knowledge and its combination with a given context---and \texttt{Quantitative/Arithmetic} questions, which require the calculation of quantities, such as probabilities or dosages. The majority of analyzed questions fall into the \texttt{Recall} category. Together with the \texttt{Reading Comprehension} category, these questions account for 95\% of annotated questions. These two categories probe the capabilities required for reading comprehension \cite{Kintsch1988}, validating the use of \qalm{} for the stated purpose of evaluating comprehension and recall. 

\begin{figure}[bh!]
\setlength\extrarowheight{-2pt}
\begin{tabular}{|p{0.95\columnwidth}|}
\hline
\textbf{Recall} \\
\hline
\textbf{Q:} During CPR, chest compressions should be delivered  at a rate of:\\
\begin{tabular}{@{}p{.35\columnwidth}p{.60\columnwidth}}
A. 80/minute.&
B. as fast as possible.\\
C. 100/minute.& 
D. varies with each patient.\\
\end{tabular} \\
\textbf{Answer:} C. 100/minute \\
\hline
\hline
 \textbf{Reading Comprehension} \\
 \hline
\textbf{Q:} A 22-year-old man comes to the physician for a routine 
health maintenance examination. He feels well. He has had a 
painless left scrotal mass since childhood. Examination 
shows a 6-cm, soft, nontender left scrotal mass that 
transilluminates; there are no bowel sounds in the 
mass. Examination of the testis shows no abnormalities. 
Which of the following is the most likely cause of 
the mass?\\

A. Accumulation of scrotal adipose tissue\newline
B. Cryptorchidism of the left testis\\
C. Dilation of the pampiniform plexus of veins around 
the testis\\
D. Persistence of a patent processus vaginalis\\
\textbf{Answer}: D. Persistence of a patent processus vaginalis \\
\hline
\hline
\textbf{Quantitative/Arithmetic} \\
\hline
\textbf{Q:} A person is prescribed Ropinirole 1.5 mg divided into three doses. How many micrograms is each dose? Choose one answer from the following:\\
\begin{tabular}{@{}p{.21\columnwidth}p{.21\columnwidth}p{.21\columnwidth}p{.21\columnwidth}}
A. 5&
B. 50&
C. 0.5&
D. 500 \\
\end{tabular} \\
\textbf{Answer:} D. 500\\
\hline
\end{tabular}
\caption{Sample questions corresponding to each category of the manual error analysis.}
\label{tab:my_label}
\end{figure}

Table~\ref{tab:manual-analysis-all} describes the accuracy of the four base and fine-tuned models: we find that \texttt{Recall} questions dominate the sample and models tend to perform best in this category, but even after fine-tuning on \qalm{}, their performance hardly surpasses 50\%, indicating that they may yet lack the necessary knowledge. Additionally, models perform worse on \texttt{Reading Comprehension} questions, suggesting that it is indeed harder to integrate necessary knowledge rather than just recalling it. Fine-tuning improves performance for all models for both types of reasoning. \texttt{Quantitative/Arithmetic} are the worst-performing category, even after fine-tuning. This is unsurprising, as arithmetic capabilities are observed to emerge with larger model scale \cite{Wei2022EmergentModels}.

\subsection{Error Analysis of LLama-2}
\label{sec:manual-analysis}
We perform a manual error analysis of the fine-tuned \llama{} (7B) model on \mcqa{}. We examine 200 non-Within Context questions where the model erred, and assign them to the \texttt{Recall}, \texttt{Reading Comprehension} and \texttt{Quantitative/Arithmetic} categories, as done previously. The model incorrectly answered 134 \texttt{Recall}, 52 \texttt{Reading Comprehension}, and 14 \texttt{Quantitative/Arithmetic} questions (Table~\ref{tab:error-categorization}). Comparing these errors to the earlier sample of 200 questions we analyze from the test set in Table \ref{tab:manual-analysis-all}, reveals that the distribution of errors for each category mirrors their general distribution in the overall test set. The prevalence of \texttt{Recall} questions in errors aligns with their dominance in medical exams like \medmcqa{}, \usmle{}, and \headqa{}. While fine-tuning on extensive medical corpora may enhance \texttt{Recall} question performance, improving on \texttt{Reading Comprehension} and \texttt{Quantitative/Arithmetic} questions might require different fine-tuning approaches, as these categories demand comprehension skills rather than mere knowledge recall.
\begin{table}[tbhp!]
\resizebox{.99\columnwidth}{!}{%

\begin{tabular}{lcc}
\hline
\textbf{Category}                 & \textbf{General Test Set} & \textbf{\llama{} Errors} \\ \hline
Recall                  & 65.5\%                          & 67\%             \\
Reading Comprehension & 29.5\%                          & 26\%          \\
Quantitative/Arithmetic  & 5\%                             & 7\%             \\
\hline
\end{tabular}
}
\caption{Distribution of Llama-2 errors across reasoning categories, compared with overall distribution of each category}
\label{tab:error-categorization}
\end{table}

%% file: section/conclusion.tex
\section{Conclusions}
In this work, we introduce \qalm{}, a comprehensive collection of clinical datasets comprising 16 multiple-choice and 6 abstractive question-answering datasets. Our study encompasses an extensive empirical investigation of open-source language models with up to 13 billion parameters. We assess their clinical and biomedical knowledge, their capacity to acquire such knowledge through training on \qalm{}, and their ability to generalize to previously unseen datasets. 

Our results highlight the strengths and limitations of \llm{s} on \mcqa{} and \aqa{}: while performing significantly better than a random guess baseline, they still fall significantly short in performance compared to proprietary language models and humans. This is true even after fine-tuning on \qalm{}, which demonstrates potential improvements, especially in the context of instruction fine-tuned models like \flan{}. 
Finally, we show inconsistencies arising from the use of different \aqa{} metrics---in future work we will supplement the automated metrics by fine-grained expert-driven manual evaluation of \llm's answers on \qalm{} to learn to automate (some dimensions of) these expert judgments.

Based on our findings, we caution on the unconstrained use of open-source \llm{s} \cite{chatdoctor,medalpaca} as assistants to help perform medical tasks or provide answers to medical queries, to experts or lay people alike, as they seem to lack the necessary medical domain knowledge.

We make the dataset, experiment code and evaluation protocol publicly available\footnote{\url{https://github.com/anand-subu/m-qalm}} to allow future developers of medical \llm{s} to assess the foundations of their models' knowledge, as our evaluation shows that  architecture of language models, the choice of datasets for pre-training and instruction fine-tuning can greatly impact their performance to the extent it can be assessed by \qalm{}.

%% file: section/appendix.tex
\appendix
\section{Datasets Used}
\label{sec:appendix-datasets}

In this section, we explain the \mcqa{} and \aqa{} datasets we used in detail. The dataset characteristics are presented in Table \ref{tab:datasets}.

\begin{enumerate}
    \item \textbf{USMLE - English}: We incorporate the \usmle{} dataset obtained from the MedQA dataset \cite{medqa}, comprising \mcqa{} questions from the Medical Licensing Exam conducted in the US. We retain this dataset's original training, validation, and test set divisions.
    
    \item \textbf{MEDMCQA}: We incorporate the \medmcqa{} dataset from \cite{medmcqa}, which comprises medical \mcqa{} from Indian Medical Entrance Exams. We retain this dataset's original training, validation, and test set splits. Similar to \citet{medpalm1}, we evaluate all models on the validation set since we do not have answers for the test set.

    \item \textbf{MMLU}: Following the design of \citet{medpalm1}, we incorporate a subset of the \mmlu{} datasets (6 datasets) \cite{mmlu} which are \mcqa{} specifically curated to assess medical domain knowledge. The subsets used are the \textbf{anatomy, clinical knowledge, college medicine, medical genetics, professional medicine} and \textbf{college biology} questions from \mmlu{}. We utilize these datasets only for evaluating models.

    \item \textbf{MEDIQA-ANS}: The MEDIQA 2019 shared task introduced the MEDIQA-QA dataset \cite{mediqa-ans} for answer-ranking, comprising consumer health questions and passages from reputable online sources. The dataset was curated by extracting passages from the text of web pages, and includes manually generated single and multi-document summaries in both extractive and abstractive forms. We employ the multi-document abstractive summary as our questions' ground truth reference answer. We specifically filter for questions and answers marked as excellent and utilize this as an \aqa{} dataset solely for evaluating models.
    
    \item \textbf{HEADQA}: We include the \headqa{} dataset \cite{headqa}, which comprises graduate-level \mcqa{} about various fields of medicine used for examinations to apply for specialization positions in the Spanish public healthcare system. We use the English version of the dataset and retain the original train, validation, and test split. 

    \item  \textbf{PubmedQA}: The PubMedQA dataset \cite{pubmedqa} is a biomedical question-answering dataset comprising 1,000 expert-annotated QA instances. Each instance necessitates reasoning over a biomedical paper's abstract to answer a relevant question. While the dataset provides long and short answers (yes, no, or maybe), we focus exclusively on the short answers for our evaluation, thereby generalizing the task as \mcqa{}. We retain the original test split of 500 questions. Additionally, we allocate 100 questions from the training set to serve as a validation set, facilitating standardized training and validation in future studies.

    \item \textbf{BioMRC}: The \biomrc{} dataset \cite{biomrc} focuses on machine reading comprehension within the biomedical domain. It is structured in a cloze-style \mcqa{} format, where questions are based on biomedical abstracts where biomedical entities are replaced with pseudo-identifiers. The task is to correctly identify the masked entity in the title from a list of masked entities. We utilize two compact versions of BioMRC: Tiny A (Setting A) and Tiny B (Setting B). The \biomrc{} dataset comprises a large training corpus, where masked entities share the same pseudo-identifier across the entire corpus. Setting A, retains the same pseudo-identifiers used for masked biomedical entities in the training corpus. This setup is beneficial when testing models trained using the BioMRC training set, allowing them to draw on previously seen patterns. Setting B, conversely, changes the pseudo-identifiers for every single question. This means that a model must rely solely on the information in the text of the question and the passage it refers to, without any help from repeated exposure to the same placeholders. While we maintain the original format for Setting B, assessing Setting A as is, is difficult as since we do not utilize the BioMRC training set, it is functionally the same as Setting B. To address this limitation, we modify Setting A to include the original entity names and their corresponding pseudo-identifiers in the answer options, based on how the original paper \cite{biomrc} assesses the performance of experts and non-experts. This aims to assess whether the model can accurately answer when provided with the information about their original entity names.
     
    \item \textbf{Processbank}: The Processbank dataset \cite{processbank} is designed for machine reading comprehension, featuring questions based on paragraphs describing biological processes. Each question, associated with a particular paragraph, has two answer options (\mcqa{}). The dataset comes with a predefined split of 435 questions (150 files) for training and 100 questions (50 files) for testing. We allocate 25 files from the training set to create a validation set while retaining the original test set for model evaluation.

    \item \textbf{QA4MRE - Alzheimer's disease QA}: The dataset proposed by Morante et al. \cite{qa4mre} contains \mcqa{} questions regarding Alzheimer's disease, aimed at assessing machine reading systems' ability to answer questions about the disease by parsing relevant documents. We have adapted this dataset as an open-ended \mcqa{} task to evaluate LLMs' ability to answer these questions based on inherent knowledge. This dataset is employed solely for model evaluation purposes.

    \item \textbf{BioASQ}: The BioASQ dataset \cite{bioasq1, bioasq2} features biomedical questions crafted by experts. We utilize the BioASQ 2022 dataset for our benchmark. The BioASQ dataset is divided into two parts: for {\mcqa{} and another for \aqa{}. For the \mcqa{} part, we filter out the yes/no questions from BioASQ, converting them into an MCQ format to create a new subset, which we term \textbf{BioASQ-MCQ}. We manually create a training-validation (train-val) split of roughly 85\%-15\% from the filtered questions, resulting in 975 training questions and 173 validation questions and retaining a test set of 123 questions. For the \aqa{} part, BioASQ provides fact, list, and bullet-type questions. We compile these into an \aqa{} dataset, ensuring a balanced representation of all question types in training and validation sets. The train-validation split results in 4733 training and 697 validation questions, with approximately 15\% of all question types in the validation set.

    \item \textbf{MASH-QA:} The MASH-QA dataset \cite{mashqa} was designed for answering medical questions based on paragraphs where answers may span multiple text segments. Initially intended for extractive answering tasks, we repurpose it as an \aqa{} task, utilizing the extractive answers as the reference ground truth.

    \item \textbf{MedQUAD:}
    The MedQUAD dataset \cite{medquad} encompasses medical question-answer pairs extracted from various National Institute of Health (NIH) websites, covering topics on diseases, drugs, and other medical entities. Only nine of the twelve websites contributing to the original dataset have answers. We segregate questions from these nine websites and devise a train-validation-test split (\aqa{}), assigning data from six websites for training, one website for validation, and two websites for testing.

    \item \textbf{TREC-2017 LiveQA}:
    We employ the TREC-2017 LiveQA dataset \cite{liveqa} for evaluation purposes. Specifically, we leverage the rankings provided within the MedQUAD evaluation process \cite{medquad} to keep question-answer pairs that have answer rating as excellent. We utilize this as an \aqa{} dataset for evaluating the model.

    \item \textbf{British Ophthalmology Practice Tests}: We employ sample questions from the Fellowship of the Royal College of Ophthalmologists (FRCOphth) exams, as provided by the Royal College of Ophthalmologists on their website \cite{opth, opth1, opth2}. These \mcqa{} questions, geared towards testing ophthalmology-related knowledge, are used for evaluation.
    
    \item \textbf{MEDINFO}: The MEDINFO dataset, introduced by Abacha et al. \cite{medinfo}, consists of real consumer questions concerning medications and drugs. It encompasses 674 question-answer pairs (\aqa{}), which we employ solely for evaluation.
    }
    \end{enumerate}

\section{Performance of other methods for \mcqa{} datasets}
\label{sec:appendix-best-perf}
We report the prior and current best scores on \mcqa{} datasets from current literature in \Cref{tab:ref_scores}. GPT-4 combined with a prompting strategy labeled MedPrompt performs the best currently on \usmle{}, \medmcqa{}, and the \mmlu{} datasets. Of the 16 datasets, we can obtain comparable scores for 12. For \headqa{}, the results reported by \cite{headqa} and \cite{multi_step_reasoning} are across individual sections, whereas we calculate the scores overall across all questions. The method proposed by \cite{multi_step_reasoning}, named \textbf{MurKe} achieves average scores of 45.5\% on Biology questions, 42.4\% on Medicine questions, 42.3\% on Nursing Questions,  48.0\% on Pharmacology questions, 44.3\% on Psychology questions and 44.3\% on Chemistry Questions, with an overall macro-average of 44.4\% across all the sections. Similarly, for the \britishopthamology{} dataset, the results reported by \cite{opth} are separate for Part 1 and Part 2 questions. Bing Chat performs the best on Part 1 questions, achieving a performance of 78.9\%, and GPT-4 with prompting obtains a performance of 88.4\% on Part 2 questions \cite{opth}. We could not find directly comparable scores for the \textbf{BioASQ} MCQ datasets as the test sets are provided in different batches, with the results on the BioASQ leaderboard also reported separately in terms of batches. We combine the questions across all the batches into one combined test set. For \biomrc{} - Tiny A, we do not have directly  comparable scores from prior works as we provide the names of the original entities along with the pseudo-identifiers to the \llm{s}, similar to how \cite{biomrc} evaluate the performance of experts and non-experts. In contrast, when evaluating the performance of systems/deep learning models, \cite{biomrc} first fine-tune models on the \biomrc{}-Lite dataset and evaluate performance on \biomrc{} - Tiny A, without providing names of the original entities to the system.

\section{Correlation between \aqa{} and \mcqa{} metrics}
\label{sec:appendix-corr}

We use \rouge{}, \bertscore{} and \meteor{} for evaluating the performance of \llm{s} for \aqa{}. We try to understand which of the three metrics might be the most reliable for evaluation. Assuming that MCQA evaluations give a more robust estimate of models’ capabilities due to the exact nature of evaluation, we calculate the correlation between the \mcqa{} accuracy and each of the \aqa{} metrics. Removing the \flan{}-ZS models as outliers, we calculate the Spearman Rank Correlation and obtain the following results:

\begin{table}[tbhp!]
\resizebox{1.\columnwidth}{!}{%

\begin{tabular}{|l|c|c|}
\hline
\textbf{Metrics}                 & \textbf{Spearman R} & \textbf{P-value} \\ \hline
\mcqa{} Accuracy and \aqa{} \rouge{}                  &   0.616                        &   0.008           \\
\mcqa{} Accuracy and \aqa{} \bertscore{} &   0.353                        &  0.164         \\
\mcqa{} Accuracy and \aqa{} \meteor{} &      -0.192                        &    0.461          \\
\hline
\end{tabular}
}
\caption{Spearman Rank Correlation between \mcqa{} accuracy and \aqa{} metrics along with their statistical significance}
\end{table}

The scores indicate that only \rouge{} scores show a reliable and statistically significant correlation to \mcqa{} Accuracy scores, suggesting that this might be the more reliable metric of the three. However, we wish to stress that these results must not be taken as definitive because the underlying assumption is that models performing better on \mcqa{} should also perform better on \aqa{}.

\section{Analysis of the causes of generalisation to unseen datasets}
\label{sec:appendix-gen-analysis}
We aim to discriminate whether \mcqa{} fine-tuned models' performance on unseen \mcqa{} datasets can be attributed to their ability to generalize in answering medical questions, or if their performance is influenced by memorization of  questions from the training set.
To this end, we examine three evaluation-only MCQ datasets not used in the training split of \qalm{}: Clinical Knowledge Tests (\mmlu{}-CK) and Medical Genetics (\mmlu{}-MG) from \mmlu{} and the \britishopthamology{} dataset.
We utilize semantic similarity algorithms to retrieve questions in the training sets that closely resemble those in these test sets and manually filter the retrieved results. 
We identify 6 out of 92, 12 out of 265, and 17 out of 100  questions in the \britishopthamology{}, \mmlu{}-CK, and \mmlu{}-MG datasets, respectively, that have similar counterparts in the \medmcqa{} dataset which was used to fine-tune the \llama{} model 
This suggests that scores might be inflated due to train-test leakage.

Next, we focus on questions that the \llama{} (7B)  model answered wrongly, but which were corrected by \mcqa{}-fine-tuning. We then cross-reference these with the closest equivalent questions in the \medmcqa{} dataset. This allows us to categorize the correct answers from near-duplicate memorization or the model's generalized learning capabilities. We find 5, 2, and 5 questions in the three investigated datasets, respectively, where the \mcqa{}-fine-tuned model outperformed its zero-shot counterpart and identified closely related questions in \medmcqa{}. Of these, 7 questions were near-duplicates with identical answers, while the remaining 5 would have required some level of clinical understanding for the model to answer them correctly.  
This suggests that the improved performance of instruction-tuned models on unseen datasets can be partially attributed to exposure to near-identical questions during training.

\begin{table*}[!t]
    \centering
    \large
    \resizebox{.99\columnwidth}{!}{%
    
    \renewcommand{\arraystretch}{1.25}
    \begin{tabular}{p{0.03cm} l | c : G G}
    &  & \multicolumn{2}{c}{\textbf{{\mcqa{}}}}  \\
    &  & Macro-Avg & Micro-Avg \\
        \hline
        \hline
        \multirow{6}{*}{\rotatebox[origin=c]{90}{\emph{Base}}} &
 \lla{} (7B)          & 31.9 & 30.7 \\
& \lla{} (13B)          & 44.1 & 
 38.9   \\ &
 \llama{} (7B)          & 42.9 & 39.6 \\
& \llama{} (13B)          & 47.1 & 43.4 \\

& \mpt{} (7B)          & 27.6 & 27.3 \\
& \falcon{} (7B)          & 34.7 & 31.6\\
\hdashline

\multirow{6}{*}{\rotatebox[origin=c]{90}{\emph{Instruction tuned}}}
& \llama{}-chat (7B)       & 45.9 & 41.2  \\
& \llama{}-chat (13B)          & 50.3 & 45.6  \\
& \mpt{}-Instruct (7B)          & 31.6 & 29.1 \\
& \falcon{}-Instruct (7B)          & 31.8 & 29.7 \\
& \flan{} (3B)          & 51.8 & 40.6 \\
& \flan{} (11B)          & 56.5 & 45.2 \\
\hdashline
       \multirow{6}{*}{\rotatebox[origin=c]{90}{\emph{~~~~~~~~~~~Finetuned}}} &

 \llama{} (7B)          & 53.5 & 52.2 \\
& \mpt{} (7B)          & 53.2 &  51.5 \\
& \falcon{} (7B)          & 49.3  & 48.6 \\
& \flan{} (3B)          & 52.9 & 47.4 \\
\hdashline

\hdashline
\multirow{3}{*}{\rotatebox[origin=c]{90}{\emph{Adapted}}}
& \chatdoctor{} (7B)      & 42.8 &  36.0   \\
& \medalpaca{} (7B)      & 48.8 & 42.3  \\
& \pmcllama{} (13B)      & 53.7 & 57.9   \\
\hline
     \end{tabular}
     }
    \caption{Micro-Average and Macro-Average Accuracies of all Models}
    \label{tab:macro_micro_results}
\end{table*}

\label{sec:appendix-all-details}
\begin{table*}[h!]
\centering
\resizebox{\textwidth}{!}{
\begin{tabular}{lcc}
\hline
\textbf{Dataset}                      & \textbf{Best Reported Score} & \textbf{Method}
\\ \hline
USMLE (4 options)            & 90.2                & GPT 4 + MedPrompt \cite{medprompt}             \\
MEDMCQA                      & 79.1                & GPT 4 + MedPrompt \cite{medprompt}           \\
PubMedQA                     & 82.0                & GPT 4 + MedPrompt \cite{medprompt}             \\
MMLU - Anatomy               & 89.6                & GPT 4 + MedPrompt \cite{medprompt}             \\
MMLU - Clinical Knowledge    & 95.8                & GPT 4 + MedPrompt \cite{medprompt}             \\
MMLU - College Biology       & 97.9                & GPT 4 + MedPrompt \cite{medprompt}             \\
MMLU - College Medicine      & 89.0                & GPT 4 + MedPrompt \cite{medprompt}             \\
MMLU - Medical Genetics      & 98.0                & GPT 4 + MedPrompt \cite{medprompt}             \\
MMLU - Professional Medicine & 95.2                & GPT 4 + MedPrompt \cite{medprompt}             \\
ProcessBank                  & 68.8                & SemanticILP (Biology Cascade) \cite{semanticilp} \\
QA4MRE                       & 55.0                  & Index Expansion  \cite{index_expansion} \cite{qa4mre}            \\
BioMRC - Tiny B              & 60.0                  & SciBERT-Max-Reader  \cite{biomrc}       
\\ \hline
\end{tabular}
}
\caption{Performance scores of various methods on various \mcqa{} datasets}
\label{tab:ref_scores}
\end{table*}

\begin{table*}[t!]
\centering
\resizebox{\textwidth}{!}{
\begin{tabular}{p{0.1cm} l c c p{11cm}}
\hline
& \textbf{Model} & \textbf{Architecture} & \textbf{\# Tokens} & \textbf{Data Source} \\
\hline
& \multicolumn{4}{l}{\emph{Base models}} \\
&\mpt{}& Decoder & 1T & Red Pajama \cite{redpajama}, The Stack \cite{thestack}, C4 \cite{t5}, mC4 \cite{mt5}, S20RC \cite{s20rc}\\
&\lla{} & Decoder & 1.4T & Common Crawl, C4 \cite{t5}, Github, Wikipedia, Gutenberg, Books3 \cite{books3}, Arxiv and Stack Exchange\\
&\falcon{}& Decoder & 1.5T & RefinedWeb \cite{refinedweb}\\
&\llama{} & Decoder & 2T & Unknown\\
\hdashline
& \multicolumn{3}{l}{\emph{Instruction tuned models}} \\
& \flan{} & Encoder-Decoder & 1T & C4 \cite{t5} and \textit{Flan-Collection} \cite{flant5} \\
& \mpt{}-Instruct  & Decoder & 1T & \mpt{}, \textit{Databricks Dolly-15k \cite{dolly}, Anthropic Helpful and Harmless \cite{rrhlf}} \\
& \falcon{}-Instruct & Decoder & 1.5T & \falcon{}, \textit{baize \cite{baize}, GPT4All, GPTeacher \footnote{https://github.com/teknium1/GPTeacher}} \\
& \llama{}-Chat & Decoder & 2T & \llama{},  \textit{Flan Collection \cite{flant5}, Private Data} \\
\hline
\end{tabular}
}
\caption{Pretrained \llm{s} considered in this paper. (Top rows) Open-source models that are decoder-only. (Bottom rows) Instruction-fine-tuned language models. 
\textbf{\# Tokens}: Number of tokens used in pretraining the model. \textbf{Data Source}: Data used for pre-training (instruction data is \emph{italicized}). 
}
\label{tab:models}
\end{table*}
\begin{table*}[ht!]
\centering
\small
\begin{tabular}{lcccc}
\hline
Parameter          & Flan-T5 XL & Llama-2 7B & Falcon 7B & MPT 7B \\ \hline
lora\_r            & 16         & 16         & 16        & 16     \\
lora\_alpha        & 16         & 16         & 16        & 16     \\
lora\_dropout      & 0.05       & 0.05       & 0.05      & 0.05   \\
bias               & none       & none       & none      & none   \\
optimizer          & adamw      & adamw      & adamw     & adamw  \\
epochs             & 4          & 4          & 4         & 4      \\
batch size         & 8          & 8          & 8         & 8      \\
model\_max\_length & 256        & 384        & 384       & 384    \\ \hline
\end{tabular}
\caption{Hyper-parameters used to train our models}
\label{tab:hyperparams}
\end{table*}

\begin{table*}[ht!]
\centering
\small
\begin{tabular}{lcc}
\hline
Parameter         & Decoder LLMs & \multicolumn{1}{c}{Encoder-Decoder LLMs} \\ \hline
Beam Size          & 3            & 3                                         \\
Repetition Penalty & 1.5          & 1.5                                       \\
Max Output Length  & 200          & 200                                       \\ \hline
\end{tabular}
\caption{Inference time parameters used for abstractive question answering}
\end{table*}

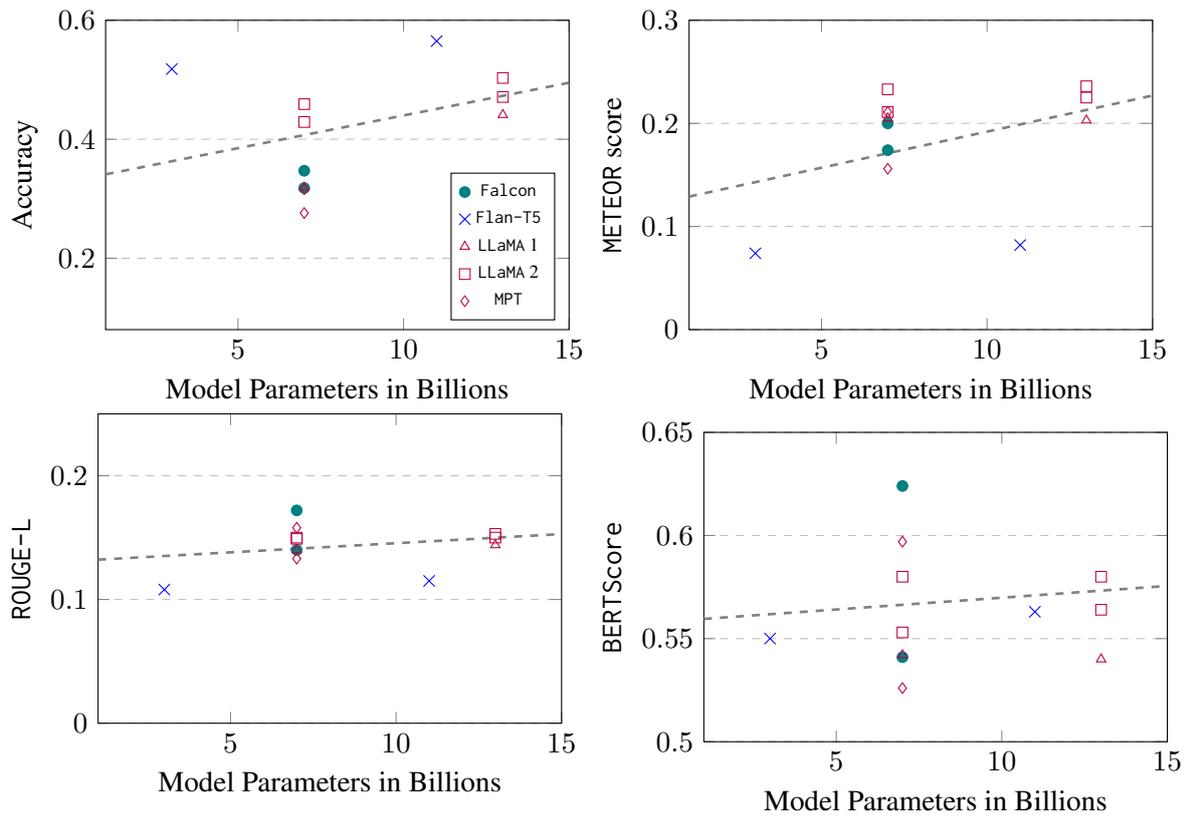
\begin{figure*}[!t]

\begin{subfigure}[t]{0.48\textwidth}
\begin{center}
\begin{tikzpicture}
\begin{axis}[
  title={},
  width=1\columnwidth,
  height=0.74\columnwidth,
  xlabel={Model Parameters in Billions},
  xmin=1, xmax=15,
  ylabel={Accuracy},
  ymin=0.08, ymax=0.6,
  legend pos=south east,
  ymajorgrids=true,
  ylabel near ticks, 
  grid style=dashed,
]
\addplot [gray, line width = 1, dashed, domain=1:15] {0.33+ 0.011* x};

\addplot+ [
    teal,
    mark options={mark=*, mark size=2pt},
    only marks,
    legend entry=\scriptsize\falcon{}
    ] 
    plot[error bars/.cd, y dir=both, y explicit] 
    coordinates {
    (7, 0.347)
    (7, 0.318)};
\addplot+ [
    blue,
    mark options={mark=x, mark size=3pt},
    only marks,
    legend entry=\scriptsize\flan{}
    ] 
    plot[error bars/.cd, y dir=both, y explicit] 
    coordinates {
    (3, 0.518)  
    (11, 0.565)
    };

\addplot+ [
    purple,
    mark=triangle,
    only marks,
    legend entry=\scriptsize\lla{}
    ] 
    plot[error bars/.cd, y dir=both, y explicit] 
    coordinates {    
    (7, 0.319)
    (13, 0.441)
    };
    
\addplot+ [
    purple,
    mark=square,
    only marks,
    legend entry=\scriptsize\llama{}
    ] 
    plot[error bars/.cd, y dir=both, y explicit] 
    coordinates {    
    (7, 0.429)
    (7, 0.459)
    (13, 0.471)
    (13, 0.503)
    };

\addplot+ [
    purple,
    mark=diamond,
    only marks,
    legend entry=\scriptsize\mpt{}
    ] 
    plot[error bars/.cd, y dir=both, y explicit] 
    coordinates {    
    (7, 0.276)
    (7, 0.316)
    };    
\end{axis}
\end{tikzpicture}
\end{center}
\end{subfigure}
\begin{subfigure}[t]{0.48\textwidth}
\begin{center}
\begin{tikzpicture}
\begin{axis}[
  title={},
  width=1\columnwidth,
  height=0.74\columnwidth,
  xlabel={Model Parameters in Billions},
  xmin=1, xmax=15,
  ylabel={\meteor{} score},
  ymin=0, ymax=0.3,
  ymajorgrids=true,
  grid style=dashed,
]
\addplot [gray, line width = 1, dashed, domain=1:15] {0.122+ 0.007* x}; 

\addplot+ [
    teal,
    mark options={mark=*, mark size=2pt},
    only marks,
    ] 
    plot[error bars/.cd, y dir=both, y explicit] 
    coordinates {
    (7, 0.2)
    (7, 0.174)};
\addplot+ [
    blue,
    mark options={mark=x, mark size=3pt},
    only marks,
    ] 
    plot[error bars/.cd, y dir=both, y explicit] 
    coordinates {
    (3, 0.074)  
    (11, 0.082)
    };

\addplot+ [
    purple,
    mark=triangle,
    only marks,
    ] 
    plot[error bars/.cd, y dir=both, y explicit] 
    coordinates {    
    (7, 0.205 )
    (13, 0.203 )
    };

\addplot+ [
    purple,
    mark=square,
    only marks,
    ] 
    plot[error bars/.cd, y dir=both, y explicit] 
    coordinates {    
    (7, 0.211)
    (7, 0.233)
    (13, 0.225)
    (13, 0.236)
    };

\addplot+ [
    purple,
    mark=diamond,
    only marks,
    ] 
    plot[error bars/.cd, y dir=both, y explicit] 
    coordinates {    
    (7, 0.211)
    (7, 0.156)
    };    
\end{axis}
\end{tikzpicture}
\end{center}

\end{subfigure}

\begin{subfigure}[t]{0.48\textwidth}
\begin{center}
\begin{tikzpicture}
\begin{axis}[
  title={},
  width=1\columnwidth,
  height=0.74\columnwidth,
  xlabel={Model Parameters in Billions},
  xmin=1, xmax=15,
  ylabel={\rouge{}},
  ymin=0, ymax=0.25,
  legend pos=south east,
  ymajorgrids=true,
  ylabel near ticks, 
  grid style=dashed,
]
\addplot [gray, line width = 1, dashed, domain=1:15] {0.1307 + 0.001478*x};
\addplot+ [
    teal,
    mark options={mark=*, mark size=2pt},
    only marks,
    ] 
    plot[error bars/.cd, y dir=both, y explicit] 
    coordinates {
    (7, 0.14)
    (7, 0.172)};
\addplot+ [
    blue,
    mark options={mark=x, mark size=3pt},
    only marks,
    ] 
    plot[error bars/.cd, y dir=both, y explicit] 
    coordinates {
    (3, 0.108)  
    (11, 0.115)
    };
\addplot+ [
    purple,
    mark=square,
    only marks,
    ] 
    plot[error bars/.cd, y dir=both, y explicit] 
    coordinates {    
    (7, 0.149)
    (7, 0.150)
    (13, 0.150)
    (13, 0.153)
    };

\addplot+ [
    purple,
    mark=triangle,
    only marks,
    ] 
    plot[error bars/.cd, y dir=both, y explicit] 
    coordinates {    
    (7, 0.140)
    (13, 0.144 )
    };
    
\addplot+ [
    purple,
    mark=diamond,
    only marks,
    ] 
    plot[error bars/.cd, y dir=both, y explicit] 
    coordinates {    
    (7, 0.133)
    (7, 0.158)
    };    
\end{axis}
\end{tikzpicture}
\end{center}
\end{subfigure}
\begin{subfigure}[t]{0.48\textwidth}
\begin{center}
\begin{tikzpicture}
\begin{axis}[
  title={},
  width=1\columnwidth,
  height=0.74\columnwidth,
  xlabel={Model Parameters in Billions},
  xmin=1, xmax=15,
  ylabel={\bertscore{}},
  ymin=0.5, ymax=0.65,
  legend pos=south east,
  ymajorgrids=true,
  grid style=dashed,
]
\addplot [gray, line width = 1, dashed, domain=1:15] { 0.5584 + 0.001143*x};
\addplot+ [
    teal,
    mark options={mark=*, mark size=2pt},
    only marks,
    ] 
    plot[error bars/.cd, y dir=both, y explicit] 
    coordinates {
    (7, 0.541)
    (7, 0.624)};
\addplot+ [
    blue,
    mark options={mark=x, mark size=3pt},
    only marks,
    ] 
    plot[error bars/.cd, y dir=both, y explicit] 
    coordinates {
    (3, 0.55)  
    (11, 0.563)
    };
\addplot+ [
    purple,
    mark=square,
    only marks,
    ] 
    plot[error bars/.cd, y dir=both, y explicit] 
    coordinates {    
    (7, 0.553)
    (7, 0.58)
    (13, 0.564)
    (13, 0.58)
    };

\addplot+ [
    purple,
    mark=triangle,
    only marks,
    ] 
    plot[error bars/.cd, y dir=both, y explicit] 
    coordinates {    
    (7, 0.542)
    (13, 0.540 )
    };

\addplot+ [
    purple,
    mark=diamond,
    only marks,
    ] 
    plot[error bars/.cd, y dir=both, y explicit] 
    coordinates {    
    (7, 0.526)
    (7, 0.597)
    };    
\end{axis}
\end{tikzpicture}
\end{center}

\end{subfigure}

\caption{Zero-shot performance of models on \mcqa{} (top-left) and \aqa{} (top-right, bottom-left and bottom-right) as a function of model size. The dashed line represents a fitted linear regression showing the correlation between the model size and the score.}
\label{fig:by-params}
\end{figure*}

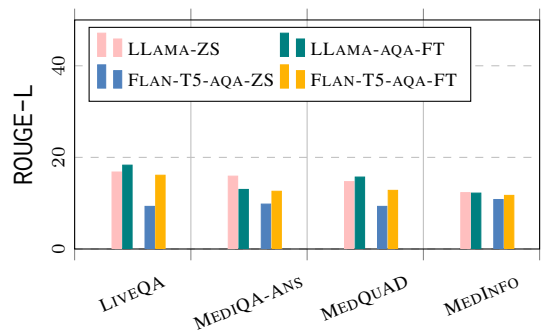
\begin{figure*}[!h]
\centering
    \begin{tikzpicture}
    \begin{axis}[
        ybar=0.5pt,
        bar width=3.5pt,
        x tick label style={rotate=23, font=\scriptsize\itshape},
        y tick label style={rotate=90, font=\scriptsize\itshape},
        height=12em,
        ylabel=\rouge{},
        width  = \columnwidth,
        x tick label as interval,
        ymajorgrids=true,
        y grid style=dashed,
        major y tick style = transparent,
        xtick={0, 1, 2, 3, 4, 5, 6, 7},
        xmajorgrids=true,
        xminorgrids=true,
        legend pos=north west,
        legend cell align={left},
        legend columns=2,
        xmin=0, xmax=4,
        ymin=0.0, ymax=50,
        xticklabels={\liveqa{}, \mediqa{}, \medquad{}, \medinfo{}}
    ]
    \addplot[legend entry=\textsc{\scriptsize LLama-ZS}, color=pink, fill=pink]  coordinates {(0.5, 16.8) (1.5, 15.9) (2.5, 14.7) (3.5, 12.3) };
    \addplot[legend entry=\textsc{\scriptsize LLama-\aqa{}-FT},, color=teal, fill=teal]  coordinates {(0.5, 18.3) (1.5, 13.0) (2.5, 15.7) (3.5, 12.2) };
    \addplot[legend entry=\textsc{\scriptsize Flan-T5-\aqa{}-ZS},, color=bblue, fill=bblue]  coordinates {(0.6, 9.3) (1.6, 9.8) (2.6, 9.3) (3.6, 10.8) };
    \addplot[legend entry=\textsc{\scriptsize Flan-T5-\aqa{}-FT},, color=rred, fill=rred]  coordinates {(0.6, 16.1) (1.6, 12.6) (2.6, 12.8) (3.6, 11.7) };

    \end{axis}
    \end{tikzpicture}
    
    \caption{Performance of base and \aqa{}-finetuned \llama{} and \flan{} models on four unseen \aqa{} test sets in terms of \rouge{}.}
    \label{fig:generalisation-aqa-rouge}
\end{figure*}

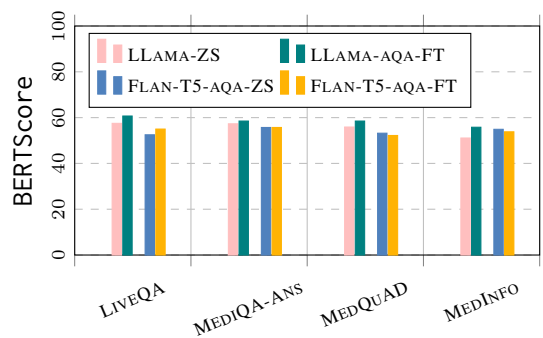
\begin{figure*}[!h]
\centering
    \begin{tikzpicture}
    \begin{axis}[
        ybar=0.5pt,
        bar width=3.5pt,
        x tick label style={rotate=23, font=\scriptsize\itshape},
        y tick label style={rotate=90, font=\scriptsize\itshape},
        height=12em,
        ylabel=\bertscore{},
        width  = \columnwidth,
        x tick label as interval,
        ymajorgrids=true,
        y grid style=dashed,
        major y tick style = transparent,
        xtick={0, 1, 2, 3, 4, 5, 6, 7},
        xmajorgrids=true,
        xminorgrids=true,
        legend pos=north west,
        legend cell align={left},
        legend columns=2,
        xmin=0, xmax=4,
        ymin=0.0, ymax=100,
        xticklabels={\liveqa{}, \mediqa{}, \medquad{}, \medinfo{}}
    ]
    \addplot[legend entry=\textsc{\scriptsize LLama-ZS}, color=pink, fill=pink]  coordinates {(0.5, 57.5) (1.5, 57.3) (2.5, 55.9) (3.5, 51.1) };
    \addplot[legend entry=\textsc{\scriptsize LLama-\aqa{}-FT},, color=teal, fill=teal]  coordinates {(0.5, 60.7) (1.5, 58.5) (2.5, 58.5) (3.5, 55.8) };
    \addplot[legend entry=\textsc{\scriptsize Flan-T5-\aqa{}-ZS},, color=bblue, fill=bblue]  coordinates {(0.6, 52.5) (1.6, 55.7) (2.6, 53.2) (3.6, 54.9) };
    \addplot[legend entry=\textsc{\scriptsize Flan-T5-\aqa{}-FT},, color=rred, fill=rred]  coordinates {(0.6, 55.0) (1.6, 55.7) (2.6, 52.2) (3.6, 53.8) };
    \end{axis}
    \end{tikzpicture}
    
    \caption{Performance of base and \aqa{}-finetuned \llama{} and \flan{} models on four unseen \aqa{} test sets in terms of \bertscore{}.}
    \label{fig:generalisation-aqa-bs}
\end{figure*}

\input{section/tables/zero_shot_mcq_scores_base}

\input{section/tables/zero_shot_ins_llms}

\input{section/tables/finetuned_on_mcq_scores}

\input{section/tables/finetuned_on_aqa_mcq_scores}

\input{section/tables/finetuned_on_mcqandaqa_mcq_scores}

\input{section/tables/existing_biomedical_llm_mcqa_scores}

\input{section/tables/rouge_scores_category}

\input{section/tables/zero_shot_aqa_base_llms}

\clearpage

\section{Prompts utilized}
In this section, we outline all prompts used for finetuning and evaluating the \llm{s}. We define \textbf{Single Context \mcqa{} Prompt} as the prompt for the \processbank{} dataset with a single paragraph context, \textbf{Multi-Context \mcqa{} Prompt} as the prompt for the \pubmedqa{} dataset with multiple paragraph contexts, \textbf{Cloze \mcqa{} Prompt} for the \biomrc{} Setting A and B datasets, \textbf{\mcqa{} Prompt} for all other \mcqa{} datasets, and \textbf{\aqa{} Prompt} for all \aqa{} datasets.

\begin{figure}[hbt!]
\centering     
\begin{tcolorbox}[title={\aqa{} Prompt for fine-tuning and evaluating \falcon{} (Base), \mpt{} (Base), \llama{} (Base) and \flan{}}]
Answer the medical question precisely and factually

Question: \{Question\}

Answer:
\end{tcolorbox}
\caption{\aqa{} prompt utilized for finetuning and evaluating these models.}
\end{figure}

\begin{figure}[ht!]
\centering     
\begin{tcolorbox}[title={\mcqa{} Prompt for fine-tuning and evaluating \falcon{} (Base), \mpt{} (Base), \llama{} (Base) and \flan{}}]
Pick the right option that answers the question

Question: \{Question\}

Options:

A. \{Option Text\}

B. \{Option Text\}

C. \{Option Text\}

D. \{Option Text\}

Answer:
\end{tcolorbox}
\caption{\mcqa{} prompt utilized for finetuning and evaluating these models.}
\end{figure}

\begin{figure}[ht!]
\centering     
\begin{tcolorbox}[title={Single Context \mcqa{} Prompt for fine-tuning and evaluating \falcon{} (Base), \mpt{} (Base), \llama{} (Base) and \flan{}}]
Given the context, pick the right choice that answers the question

Context: \{Context Paragraph\}

Question: \{Question\}

Options:

A. \{Option Text\}

B. \{Option Text\}

Answer:
\end{tcolorbox}
\caption{Single Context \mcqa{} prompt utilized for finetuning and evaluating these models on the \processbank{} dataset.}
\end{figure}

\begin{figure}[hbt!]
\centering     
\begin{tcolorbox}[title={Multi Context \mcqa{} Prompt for fine-tuning and evaluating \falcon{} (Base), \mpt{} (Base), \llama{} (Base) and \flan{}}]
Given the context, pick the right choice that answers the question

Contexts: \{Context Paragraph 1\}

\{Context Paragraph 2\}

...

\{Context Paragraph N\}

Question: \{Question\}

Options:

A. \{Option Text\}

B. \{Option Text\}

C. \{Option Text\}

Answer:
\end{tcolorbox}
\caption{Multi Context \mcqa{} prompt utilized for finetuning and evaluating these models on the\pubmedqa{} dataset.}
\end{figure}

\begin{figure}[hbt!]
\centering     
\begin{tcolorbox}[title={Cloze \mcqa{} Prompt for evaluating fine-tuned \falcon{} (Base), \mpt{} (Base), \llama{} (Base) and \flan{}}]
Given the context, pick the right choice that corresponds to the XXXX in the question

Context: \{Context Paragraph\}

Question: \{Question\}

Options:

A. \{Option Text\}

B. \{Option Text\}

C. \{Option Text\}

Answer:
\end{tcolorbox}
\caption{Cloze \mcqa{} prompt utilized for evaluating the fine-tuned models on the \biomrc{} datasets.}
\end{figure}

\begin{figure}[htb!]
\centering     
\begin{tcolorbox}[title={\mcqa{} Prompt for evaluating \falcon{} (Base and Instruct), \mpt{} (Base), \lla{} (Base), \llama{} (Base) and \flan{} without any fine-tuning}]
Pick the right option that answers the question

Question: \{Example 1\}

Options:

A. \{Option Text\}

B. \{Option Text\}

C. \{Option Text\}

D. \{Option Text\}

Answer:\{Correct Option\}

...

Question: \{Example K\}

Options:

A. \{Option Text\}

B. \{Option Text\}

C. \{Option Text\}

D. \{Option Text\}

Answer:\{Correct Option\}

Question: \{Question\}

Options:

A. \{Option Text\}

B. \{Option Text\}

C. \{Option Text\}

D. \{Option Text\}

Answer:
\end{tcolorbox}
\caption{\mcqa{} prompt utilized for evaluating models prior to any fine-tuning. 5-shot prompting is utilized for the MMLU datasets whereas 1-shot prompting is utilized for all other \mcqa{} datasets when evaluating non-finetuned models.}
\end{figure}

\begin{figure}[ht!]
\centering     
\begin{tcolorbox}[title={\aqa{} Prompt for evaluating \falcon{} (Base and Instruct), \mpt{} (Base), \lla{} (Base), \llama{} (Base) and \flan{} without any fine-tuning}]
Answer the medical question precisely and factually

Question: \{Question\}

Answer:
\end{tcolorbox}
\caption{\aqa{} prompt utilized for evaluating the models.}
\end{figure}

\begin{figure}[hbt!]
\centering     
\begin{tcolorbox}[title={Single Context \mcqa{} Prompt for evaluating \falcon{} (Base and Instruct), \mpt{} (Base), \lla{} (Base), \llama{} (Base) and \flan{} without any fine-tuning}]
Given the context, pick the right choice that answers the question

Context: \{Context Paragraph\}

Question: \{Example Question\}

Options:

A. \{Option Text\}

B. \{Option Text\}

Answer:\{Correct Option\}

Context: \{Context Paragraph\}

Question: \{Question\}

Options:

A. \{Option Text\}

B. \{Option Text\}

Answer:
\end{tcolorbox}
\caption{Single Context \mcqa{} prompt utilized for evaluating the \processbank{} dataset.}
\end{figure}

\begin{figure}[hbt!]
\centering     
\begin{tcolorbox}[title={Cloze \mcqa{} Prompt for evaluating \falcon{} (Base and Instruct), \mpt{} (Base), \lla{} (Base), \llama{} (Base) and \flan{} without any fine-tuning}]
Given the context, pick the right choice that corresponds to the XXXX in the question

Context: \{Context Paragraph\}

Question: \{Example Question\}

Options:

A. \{Option Text\}

B. \{Option Text\}

C. \{Option Text\}

Answer:\{Correct Option\}

Context: \{Context Paragraph\}

Question: \{Question\}

Options:

A. \{Option Text\}

B. \{Option Text\}

C. \{Option Text\}

Answer:
\end{tcolorbox}
\caption{Cloze \mcqa{} prompt utilized for evaluating the \biomrc{} datasets in settings
A and B.
.}
\end{figure}

\begin{figure}[htb!]
\centering     
\begin{tcolorbox}[title={Multi Context \mcqa{} Prompt for evaluating \falcon{} (Base and Instruct), \mpt{} (Base), \lla{} (Base), \llama{} (Base) and \flan{} without any fine-tuning}]
Given the context, pick the right choice that answers the question

Contexts: \{Context Paragraph 1\}

\{Context Paragraph 2\}

...

\{Context Paragraph N\}

Question: \{Example Question\}

Options:

A. \{Option Text\}

B. \{Option Text\}

C. \{Option Text\}

Answer:\{Correct Option\}

Contexts: \{Context Paragraph 1\}

\{Context Paragraph 2\}

...

\{Context Paragraph N\}

Question: \{Question\}

Options:

A. \{Option Text\}

B. \{Option Text\}

C. \{Option Text\}

Answer:
\end{tcolorbox}
\caption{Multi-Context \mcqa{} prompt utilized for evaluating the \pubmedqa{} dataset.}
\end{figure}

\begin{figure}[htb!]
\centering     
\begin{tcolorbox}[title={\mcqa{} Prompt for evaluating \llama{} (Chat) Models without any fine-tuning}]
[INST] <<SYS>>

Pick the right option that answers the question
<</SYS>>\newline

Question: \{Example Question\}

Options:

A. \{Option Text\}

B. \{Option Text\}

C. \{Option Text\}

D. \{Option Text\} [/INST] Answer:\{Correct Option\} </s><s>[INST] Question: \{Question\}

Options:

A. \{Option Text\}

B. \{Option Text\}

C. \{Option Text\}

D. \{Option Text\} [/INST] Answer:
\end{tcolorbox}
\caption{\mcqa{} prompt utilized foe evaluating the model. 5-shot prompting is utilized for the MMLU datasets whereas 1-shot prompting is utilized for all other \mcqa{} datasets.}
\end{figure}

\begin{figure}[htb!]
\centering     
\begin{tcolorbox}[title={Single Context \mcqa{} Prompt for evaluating \llama{} (Chat) Models without any fine-tuning}]
[INST] <<SYS>>

Given the context, pick the right choice that answers the question

<</SYS>>\newline

Context: \{Context Paragraph\}

Question: \{Example Question\}

Options:

A. \{Option Text\}

B. \{Option Text\} [/INST] Answer:\{Correct Option\} </s><s>[INST] 
Context: \{Context Paragraph\}

Question: \{Question\}

Options:

A. \{Option Text\}

B. \{Option Text\} [/INST] Answer:
\end{tcolorbox}
\caption{Single Context \mcqa{} prompt utilized for evaluating models on the \processbank{} dataset.}
\end{figure}

\begin{figure}[htb!]
\centering     
\begin{tcolorbox}[title={\aqa{} Prompt for evaluating \llama{} (Chat) Models without any fine-tuning}]
[INST] <<SYS>>

Answer the medical question precisely and factually

<</SYS>>\newline

Question: \{Question\} [/INST]
\end{tcolorbox}
\caption{\aqa{} prompt utilized for evaluating the model}
\end{figure}

\begin{figure}[htb!]
\centering     
\begin{tcolorbox}[title={Cloze \mcqa{} Prompt for evaluating \llama{} (Chat) Models without any fine-tuning}]
[INST] <<SYS>>

Given the context, pick the right choice that corresponds to the XXXX in the question

<</SYS>>\newline

Context: \{Context Paragraph\}

Question: \{Example Question\}

Options:

A. \{Option Text\}

B. \{Option Text\} [/INST] Answer:\{Correct Option\} </s><s>[INST] Context: \{Context Paragraph\}

Question: \{Question\}

Options:

A. \{Option Text\}

B. \{Option Text\} [/INST] Answer:
\end{tcolorbox}
\caption{Cloze \mcqa{} prompt utilized for evaluating models on the \biomrc{} datasets in settings A and B.}
\end{figure}

\begin{figure*}[htb!]
\centering     
\begin{tcolorbox}[title={Multi Context \mcqa{} Prompt for evaluating \llama{} (Chat) Models without any fine-tuning}]
[INST] <<SYS>>

Given the context, pick the right choice that answers the question

<</SYS>>\newline

Contexts: \{Context Paragraph 1\}

\{Context Paragraph 2\}

...

\{Context Paragraph N\}

Question: \{Example Question\}

Options:

A. \{Option Text\}

B. \{Option Text\}

C. \{Option Text\} [/INST] Answer:\{Correct Option\} </s><s>[INST] Contexts: \{Context Paragraph 1\}

\{Context Paragraph 2\}

...

\{Context Paragraph N\}

Question: \{Question\}

Options:

A. \{Option Text\}

B. \{Option Text\} 

C. \{Option Text\} [/INST] Answer:
\end{tcolorbox}
\caption{Multi-Context \mcqa{} prompt utilized for evaluating models on the \pubmedqa{} dataset.}
\end{figure*}

\begin{figure*}[htb!]
\centering     
\begin{tcolorbox}[title={\aqa{} Prompt for evaluating \mpt{} Instruct without fine-tuning}]
Below is an instruction that describes a task. Write a response that appropriately completes the request.

\#\#\# Instruction:

Answer the medical question precisely and factually. Question: \{Question\}

\#\#\# Response:

Answer:
\end{tcolorbox}
\caption{\aqa{} prompt utilized for evaluating the model.}
\end{figure*}

\begin{figure*}[htb!]
\centering     
\begin{tcolorbox}[title={\mcqa{} Prompt for evaluating \mpt{} Instruct without fine-tuning}]
Below is an instruction that describes a task. Write a response that appropriately completes the request.

\#\#\# Instruction:

Pick the right option that answers the question. Question: \{Example Question 1\}

Options:

A. \{Option Text\}

B. \{Option Text\}

C. \{Option Text\}

D. \{Option Text\}

\#\#\# Response:

Answer:\{Correct Option\}

...

\#\#\# Instruction:

Pick the right option that answers the question. Question: \{Example Question K\}

Options:

A. \{Option Text\}

B. \{Option Text\}

C. \{Option Text\}

D. \{Option Text\}

\#\#\# Response:

Answer:\{Correct Option\}

\#\#\# Instruction:

Pick the right option that answers the question. Question: \{Question\}

Options:

A. \{Option Text\}

B. \{Option Text\}

C. \{Option Text\}

D. \{Option Text\}

\#\#\# Response:

Answer:
\end{tcolorbox}
\caption{\mcqa{} prompt utilized for evaluating the model. 5-shot prompting is utilized for the MMLU datasets whereas 1-shot prompting is utilized for all other \mcqa{} datasets.}
\end{figure*}

\begin{figure*}[htb!]
\centering     
\begin{tcolorbox}[title={Single Context \mcqa{} Prompt for evaluating \mpt{} Instruct without fine-tuning}]
Below is an instruction that describes a task. Write a response that appropriately completes the request.

\#\#\# Instruction:

Given the context, pick the right choice that answers the question. Context: \{Context Paragraph\}

Question: \{Example Question\}

Options:

A. \{Option Text\}

B. \{Option Text\}

\#\#\# Response:

Answer:\{Correct Option\}

\#\#\# Instruction:

Given the context, pick the right choice that answers the question. Context: \{Context Paragraph\}

Question: \{Question\}

Options:

A. \{Option Text\}

B. \{Option Text\}

\#\#\# Response:

Answer:
\end{tcolorbox}
\caption{Single Context \mcqa{} prompt utilized for evaluating the model on the \processbank{} dataset.}
\end{figure*}

\begin{figure*}[htb!]
\centering     
\begin{tcolorbox}[title={Multi Context \mcqa{} Prompt for evaluating \mpt{} Instruct without fine-tuning}]
Below is an instruction that describes a task. Write a response that appropriately completes the request.

\#\#\# Instruction:

Given the contexts, pick the right choice that answers the question. Contexts: \{Context Paragraph 1\}

\{Context Paragraph 2\}

...

\{Context Paragraph N\}

Question: \{Example Question 1\}

Options:

A. \{Option Text\}

B. \{Option Text\}

C. \{Option Text\}

\#\#\# Response:

Answer:\{Correct Option\}

\#\#\# Instruction:

Given the contexts, pick the right choice that answers the question. Contexts: \{Context Paragraph 1\}

\{Context Paragraph 2\}

...

\{Context Paragraph N\}

Question: \{Question\}

Options:

A. \{Option Text\}

B. \{Option Text\}

C. \{Option Text\}

\#\#\# Response:

Answer:
\end{tcolorbox}
\caption{Multi-Context \mcqa{} prompt utilized for evaluating the model on the \pubmedqa{} dataset.}
\end{figure*}

\begin{figure*}[htb!]
\centering     
\begin{tcolorbox}[title={Cloze \mcqa{} Prompt for evaluating \mpt{} Instruct without fine-tuning}]
Below is an instruction that describes a task. Write a response that appropriately completes the request.

\#\#\# Instruction:

Given the context, pick the right choice that corresponds to the XXXX in the question. Context: \{Context Paragraph\}

Question: \{Example Question\}

Options:

A. \{Option Text\}

B. \{Option Text\}

\#\#\# Response:

Answer:\{Correct Option\}

\#\#\# Instruction:

Given the context, pick the right choice that corresponds to the XXXX in the question. Context: \{Context Paragraph\}

Question: \{Question\}

Options:

A. \{Option Text\}

B. \{Option Text\}

\#\#\# Response:

Answer:
\end{tcolorbox}
\caption{Cloze \mcqa{} prompt utilized for evaluating the model on the \biomrc{} datasets in settings A and B.}
\end{figure*}

\input{section/prompts/chatdoctor_prompts}

\input{section/prompts/medalpaca_prompts}

\input{section/prompts/pmc_llama_prompts}

\label{sec:appendix}

%% file: section/tables/zero_shot_mcq_scores_base.tex
\begin{table*}[ht!]
\centering
\resizebox{1.8\columnwidth}{!}{%

\begin{tabular}{lccccccc}
\hline
Dataset           & Random Baseline           &  \falcon{} (7B) & \mpt{} (7B) &  \llama{} (7B) & \llama{} (13B) & \lla{} (7B) & \lla{} (13B)  \\
\hline
\bioasqmcq{}     &  50.0             &  72.4          & 33.3       & 67.5           & 35.8    & 35.0 & 37.4        \\
\biomrc{} Tiny A  &  21.6             & 26.7          & 23.3       & 30.0           & 53.3 & 26.7 & 60.0           \\
\biomrc{} Tiny B   &  18.1                & 16.7          & 13.3       & 26.7           & 20.0   & 13.3 & 33.3         \\
\mmlu{} - Anatomy   &  25.0          &  28.1          & 26.7       & 40.7           & 54.1  & 37.8 & 45.9          \\
\mmlu{} - Clinical Knowledge & 25.0        & 32.5          & 29.8       & 38.1           & 57.7   & 35.5 & 43.4         \\
\mmlu{} - College Biology    & 25.0        & 27.1          & 22.2       & 39.6           & 58.3  & 35.4 & 44.4          \\
\mmlu{} - College Medicine   & 25.0       & 30.6          & 26.6       & 35.3           & 54.3  & 25.4 & 42.2          \\
\mmlu{} - Medical Genetics    & 25.0      & 33.0          & 27.0       & 49.0           & 52.0  & 34.0 & 42.0          \\
\mmlu{} - Professional Medicine & 25.0     & 44.1          & 20.2       & 44.1           & 53.7  & 28.3 & 47.1          \\
\headqa{}      &   25.0                   & 27.8          & 28.0       & 40.4           & 48.5  & 34.4 & 40.6          \\
\medmcqa{}      &  25.0                   & 30.4          & 26.5       & 36.0           & 37.5  & 27.0 & 35.9          \\
\britishopthamology{}  &  25.0                    & 21.7          & 28.3       & 27.2           & 30.4 & 20.7 & 39.1            \\
\processbank{}         & 50.0              & 50.7          & 56.0       & 75.3           & 83.3  & 63.3 & 74.0          \\
\pubmedqa{}             & 33.3            & 57.0          & 33.8       & 60.4           & 33.8  & 34.2 & 34.8          \\
\qamre{}                 & 20.0          & 30.0          & 22.5       & 40.0           & 37.5  & 30.0 & 47.5          \\
\usmle{}                  & 25.0           & 27.0          & 24.2       & 35.3           & 42.9  & 29.1 & 37.5          \\
\hline
Average                    & 27.7      & 34.7          & 27.6       & 42.9           & 47.1  & 31.9 & 44.1 \\
\hline
\end{tabular}
}
\caption{\mcqa{} scores of \llm{s} in the zero-shot setting along with a random baseline. When calculating the random baselines for each dataset, for datasets with the same number of options for all questions, we set the score as the reciprocal of the number of options. For datasets with variable number of options per question, we calculate the score for each question as the reciprocal of the number of options for that question and then average all values. We utilize 5-shot prompting for the \mmlu{} datasets and 1-shot prompting for other datasets to evaluate the models.}
\label{tab:mcqa_zs}
\end{table*}

%% file: section/tables/zero_shot_ins_llms.tex
\begin{table*}[ht!]
\centering
\resizebox{1.8\columnwidth}{!}{%
\begin{tabular}{lcccccc}
\hline
\textbf{Dataset}                   & \flan{} (3B) & \falcon{} (7B) & \mpt{} (7B) & \llama{} (7B) Chat & \flan{} (11B) & \llama{} (13B) Chat \\ \hline
\bioasqmcq{}                  & 43.9                     & 45.5                             & 34.1                          & 69.9                          & 48.8                      & 65.0                           \\
\biomrc{} Tiny A                      & 73.3                     & 30.0                             & 23.3                          & 26.7                          & 63.3                      & 33.3                           \\
\biomrc{} Tiny B                      & 46.7                     & 23.3                             & 23.3                          & 20.0                          & 60.0                      & 26.7                           \\
\mmlu{} - Anatomy                & 46.7                     & 27.4                             & 32.6                          & 44.4                          & 48.9                      & 52.6                           \\
\mmlu{} - Clinical Knowledge    & 52.1                     & 31.7                             & 36.6                          & 54.3                          & 61.9                      & 57.7                           \\
\mmlu{} - College Biology       & 48.6                     & 25.0                             & 29.9                          & 55.6                          & 54.9                      & 59.0                           \\
\mmlu{} - College Medicine      & 41.6                     & 27.7                             & 30.1                          & 44.5                          & 52.6                      & 46.2                           \\
\mmlu{} - Medical Genetics      & 50.0                     & 32.0                             & 32.0                          & 60.0                          & 55.0                      & 56.0                           \\
\mmlu{} - Professional Medicine & 42.6                     & 37.9                             & 28.3                          & 45.2                          & 55.1                      & 51.1                           \\
\headqa{}                       & 42.9                     & 26.1                             & 30.2                          & 43.9                          & 49.1                      & 51.3                           \\
\medmcqa{}                       & 33.1                     & 29.8                             & 27.2                          & 35.0                          & 36.4                      & 39.3                           \\
\britishopthamology{}                 & 26.1                     & 32.6                             & 30.4                          & 26.1                          & 25.0                      & 27.2                           \\
\processbank{}                  & 93.3                     & 52.0                             & 56.7                          & 72.0                          & 95.3                      & 80.0                           \\
\pubmedqa{}                     & 70.0                     & 47.4                             & 35.6                          & 61.6                          & 70.8                      & 45.2                           \\
\qamre{}                       & 82.5                     & 15.0                             & 30.0                          & 40.0                          & 87.5                      & 72.5                           \\
\usmle{}                      & 36.1                     & 25.1                             & 24.6                          & 35.6                          & 39.7                      & 42.2                           \\ \hline
Average                            & 51.8                     & 31.8                             & 31.6                          & 45.9                          & 56.5                      & 50.3  \\ \hline                        
\end{tabular}
}
\caption{\mcqa{} scores of Instruction-tuned \llm{s} in the zero-shot setting. We utilize 5-shot prompting for the \mmlu{} datasets and 1-shot prompting for other datasets to evaluate these models.}
\label{tab:mcqa_is}
\end{table*}

%% file: section/tables/finetuned_on_mcq_scores.tex
\begin{table*}[ht!]
\centering
\resizebox{1.8\columnwidth}{!}{
\begin{tabular}{lcccc}
\hline
\textbf{Dataset}                   & \flan{} (3B) & \falcon{} (7B) & \mpt{} (7B) & \llama{} (7B) \\ \hline
\bioasqmcq{}                  & 73.2                      & 80.5                        & 78.9                     & 81.3                         \\
\biomrc{} Tiny A                      & 53.3                      & 23.3                        & 26.7                     & 23.3                         \\
\biomrc{} Tiny B                     & 26.7                      & 23.3                        & 20.0                     & 26.7                         \\
\mmlu{} - Anatomy                & 43.7                      & 43.7                        & 45.9                     & 54.1                         \\
\mmlu{} - Clinical Knowledge    & 54.0                      & 52.8                        & 53.2                     & 59.6                         \\
\mmlu{} - College Biology       & 47.2                      & 46.5                        & 56.9                     & 61.1                         \\
\mmlu{} - College Medicine      & 44.5                      & 53.2                        & 50.3                     & 52.0                         \\
\mmlu{} - Medical Genetics      & 47.0                      & 55.0                        & 60.0                     & 62.0                         \\
\mmlu{} - Professional Medicine & 48.5                      & 50.0                        & 49.3                     & 59.6                         \\
\headqa{}                       & 49.0                      & 47.7                        & 52.4                     & 53.9                         \\
\medmcqa{}                       & 43.0                      & 45.9                        & 48.4                     & 48.3                         \\
\britishopthamology{}                 & 34.8                      & 30.4                        & 35.9                     & 31.5                         \\
\processbank{}                 & 92.7                      & 69.3                        & 84.7                     & 75.3                         \\
\pubmedqa{}                     & 74.2                      & 70.8                        & 73.4                     & 70.6                         \\
\qamre{}                    & 75.0                      & 50.0                        & 70.0                     & 50.0                         \\
\usmle{}                        & 39.7                      & 46.3                        & 45.7                     & 46.1                         \\ \hline
Average                            & 52.9                      & 49.3                        & 53.2                     & 53.5                        \\ \hline
\end{tabular}
}
\caption{\mcqa{} scores of \llm{s} finetuned with QLora on \mcqa{} datasets from the \qalm{} benchmark. We evaluate these models without any examples in the prompt.}
\label{tab:mcqa_ft}
\end{table*}

%% file: section/tables/finetuned_on_aqa_mcq_scores.tex
\begin{table*}[ht!]
\centering
\resizebox{1.8\columnwidth}{!}{

\begin{tabular}{lcccc}
\hline
\textbf{Dataset}                   & \flan{} (3B) & \falcon{} (7B) & \mpt{} (7B) & \llama{} (7B) \\ \hline
\bioasqmcq{}                  & 0.8                      & 13.8                       & 14.6                    & 7.3                         \\
\biomrc{} Tiny A                      & 50.0                     & 23.3                       & 10.0                    & 16.7                        \\
\biomrc{} Tiny B                      & 36.7                     & 23.3                       & 16.7                    & 16.7                        \\
\mmlu{} - Anatomy                 & 43.0                     & 24.4                       & 34.8                    & 38.5                        \\
\mmlu{} - Clinical Knowledge    & 50.9                     & 25.3                       & 28.7                    & 40.8                        \\
\mmlu{} - College Biology         & 42.4                     & 23.6                       & 34.7                    & 38.9                        \\
\mmlu{} - College Medicine      & 41.0                     & 27.2                       & 26.0                    & 37.6                        \\
\mmlu{} - Medical Genetics      & 45.0                     & 31.0                       & 22.0                    & 49.0                        \\
\mmlu{} - Professional Medicine & 41.2                     & 44.1                       & 18.4                    & 46.7                        \\
\headqa{}                        & 38.7                     & 21.5                       & 24.8                    & 31.1                        \\
\medmcqa{}                       & 27.0                     & 21.7                       & 20.2                    & 23.0                        \\
\britishopthamology{}                  & 22.8                     & 23.9                       & 16.3                    & 19.6                        \\
\processbank{}                   & 88.0                     & 54.7                       & 42.0                    & 50.7                        \\
\pubmedqa{}                      & 67.2                     & 57.2                       & 54.6                    & 47.8                        \\
\qamre{}                       & 77.5                     & 35.0                       & 10.0                    & 15.0                        \\
\usmle{}                        & 34.2                     & 22.9                       & 23.9                    & 22.9                        \\ \hline
Average                            & 44.1                     & 29.6                       & 24.9                    & 31.4                 \\ \hline      
\end{tabular}
}
\caption{\mcqa{} scores of \llm{s} finetuned with QLora on \aqa{} datasets only from the \qalm{} benchmark. We utilize 5-shot prompting for the \mmlu{} datasets and 1-shot prompting for other datasets to evaluate these models.}
\label{tab:mcqa_ft_aqa}
\end{table*}

%% file: section/tables/finetuned_on_mcqandaqa_mcq_scores.tex
\begin{table*}[ht!]
\centering
\resizebox{1.8\columnwidth}{!}{
\begin{tabular}{lcccc}
\hline
\textbf{Dataset}                   & \flan{} (3B) & \falcon{} (7B) & \mpt{} (7B) & \llama{} (7B) \\ \hline
\bioasqmcq{}                 & 71.5                      & 80.5                        & 79.7                     & 79.7                         \\
\biomrc{} Tiny A                      & 50.0                      & 43.3                        & 36.7                     & 26.7                         \\
\biomrc{} Tiny B                     & 30.0                      & 6.7                         & 20.0                     & 26.7                         \\
\mmlu{} - Anatomy                  & 40.7                      & 45.2                        & 47.4                     & 52.6                         \\
\mmlu{} - Clinical Knowledge    & 51.7                      & 52.5                        & 50.9                     & 55.5                         \\
\mmlu{} - College Biology        & 43.8                      & 51.4                        & 57.6                     & 61.1                         \\
\mmlu{} - College Medicine      & 41.6                      & 48.0                        & 54.3                     & 52.6                         \\
\mmlu{} - Medical Genetics      & 52.0                      & 59.0                        & 55.0                     & 65.0                         \\
\mmlu{} - Professional Medicine & 47.1                      & 46.0                        & 50.4                     & 59.9                         \\
\headqa{}                      & 47.5                      & 47.4                        & 51.2                     & 54.2                         \\
\medmcqa{}                       & 41.7                      & 45.2                        & 47.4                     & 48.0                         \\
\britishopthamology{}                   & 32.6                      & 28.3                        & 38.0                     & 28.3                         \\
\processbank{}                 & 91.3                      & 73.3                        & 79.3                     & 83.3                         \\
\pubmedqa{}                    & 71.4                      & 67.8                        & 72.8                     & 71.8                         \\
\qamre{}                      & 72.5                      & 52.5                        & 60.0                     & 67.5                         \\
\usmle{}                        & 40.9                      & 45.7                        & 44.3                     & 45.6                         \\ \hline
Average                            & 51.7                      & 49.5                        & 52.8                     & 54.9 \\ \hline                       
\end{tabular}
}
\caption{\mcqa{} scores of \llm{s} finetuned with QLora on both \mcqa{} and \aqa{} data from the \qalm{} benchmark. We evaluate these models without any examples in the prompt.}
\label{tab:mcqa_ft_all}
\end{table*}

%% file: section/tables/existing_biomedical_llm_mcqa_scores.tex
\begin{table*}[ht!]
\centering
\resizebox{1.8\columnwidth}{!}{
\begin{tabular}{lccc}
\hline
Dataset                            & \chatdoctor{} (7B) & \medalpaca{} (7B) & \pmcllama{} (13B) \\ \hline
\bioasqmcq{}                  & 65.0      & 50.4        & 13.0         \\
\biomrc{} Tiny A                      & 20.0      & 16.7        & 30.0         \\
\biomrc{} Tiny B                      & 36.7      & 23.3        & 16.7         \\
\mmlu{} - Anatomy                & 43.7     & 60.0        & 63.0         \\
\mmlu{} - Clinical Knowledge    & 43.4      & 60.0        & 62.3         \\
\mmlu{} - College Biology       & 39.6      & 64.6        & 64.6         \\
\mmlu{} - College Medicine      & 32.4      & 52.6        & 53.2         \\
\mmlu{} - Medical Genetics      & 55.0      & 69.0        & 70.0         \\
\mmlu{} - Professional Medicine & 47.1      & 67.3        & 67.6         \\
\headqa{}                       & 37.2      & 45.1        & 59.1         \\
\medmcqa{}                      & 29.4      & 35.0        & 56.5         \\
\britishopthamology{}                  & 30.4       & 23.9        & 46.7         \\
\processbank{}                  & 62.0      & 67.3        & 74.7         \\
\pubmedqa{}                     & 67.4      & 40.8        & 72.6         \\
\qamre{}                        & 45.0      & 62.5        & 55.0         \\
\usmle{}                        & 31.3      & 42.4        & 54.7         \\ \hline
Average                         & 42.8      & 48.8        & 53.7  \\
\hline
\end{tabular}
}
\caption{\mcqa{} scores of \chatdoctor{} (7B) , \medalpaca{} (7B) and \pmcllama{} (13B). To evaluate \chatdoctor{}, we utilize 5-shot prompting for the \mmlu{} datasets and 1-shot prompting for other datasets to evaluate these models. We evaluate \medalpaca{} (7B) and \pmcllama{} (13B) directly without any examples in the prompt.}
\label{tab:biomed_pretrained_llms}
\end{table*}

%% file: section/tables/rouge_scores_category.tex
\begin{table*}[ht!]
    \centering
    \resizebox{2.2\columnwidth}{!}{
    \begin{tabular}{l c c c c c c c c c c}
 & \\ \hline
    Category & Support & \flan{} (ZS) & \flan{} (FT) & \mpt{} (ZS) & \mpt{} (FT) & \falcon{} (ZS) & \falcon{} (FT) & \llama{} (ZS) & \llama{} (FT) \\
        \hline
        \hline
Consumer Health Dataset Questions                          &  1449        &  10.5      &  13.4          &   12.6                        & 14.6         & 13.2         &  14.6          &  13.7                             & 14.5          \\
General Biomedical Dataset Questions                          &  363        &  15.0       & 26.6              & 11,4                           & 28.9         &  13.9        & 27.8           &  15.8                            &  30.0         \\
General Medical Dataset Questions                          &  200        &  9.3       & 12.8              & 13.7                           & 14.0         &  14.3        & 14.8           &  14.7                            &  15.7         \\

\hline
     \end{tabular}
         }
\caption{Performance of \llm{s} in the zero-shot and fine-tuned setting across various categories across various dataset categories in terms of Rouge Score}

\label{tab:results-rouge-categories}

\end{table*}

%% file: section/tables/zero_shot_aqa_base_llms.tex
\begin{table*}[ht!]
\centering
\resizebox{2.1\columnwidth}{!}{
\begin{tabular}{lcccccccccccccccccclll}
\hline
Model         & \multicolumn{3}{c}{\bioasqqa{}} & \multicolumn{3}{c}{\liveqa{}} & \multicolumn{3}{c}{\mashqa{}} & \multicolumn{3}{c}{\medinfo{}} & \multicolumn{3}{c}{\mediqa{}} & \multicolumn{3}{c}{\medquad{}} & \multicolumn{3}{c}{Average} \\ \hline
              & RL      & BS      & MTR    & RL      & BS      & MTR    & RL      & BS      & MTR    & RL      & BS      & MTR     & RL      & BS      & MTR    & RL      & BS      & MTR     & RL      & BS      & MTR     \\ \hline
\falcon{} (7B)   & 13.9    & 53.1    & 22.5   & 15.4    & 55.8    & 17.4   & 13.4    & 53.7    & 22.0   & 12.1    & 51.1    & 17.8    & 15.3    & 56.1    & 21.7   & 14.3    & 54.7    & 18.4    & 14.0    & 54.1    & 20.0    \\
\mpt{} (7B)      & 11.4    & 50.1    & 21.7   & 15.7    & 55.2    & 20.9   & 12.8    & 52.3    & 23.0   & 11.2    & 49.6    & 18.4    & 14.8    & 55.6    & 23.3   & 13.7    & 53.2    & 19.4    & 13.3    & 52.6    & 21.1    \\
\lla{} (7B)  &  13.8   & 53.4    & 23.3   & 15.4    & 55.8    & 18.9   & 13.5    & 54.1    & 22.2   & 11.6    & 51.4    & 17.9    & 15.5    & 56.8    & 22.5   & 14.3    & 54.0    & 18.5    & 14.0    & 54.2    & 20.5    \\
\lla{} (13B)  &  14.6   & 53.3    & 22.8   & 16.7    & 55.7    & 19.7   & 13.1    & 53.3    & 20.9   & 12.5    & 51.7    & 18.6    & 15.4    & 57.0    & 22.1   & 14.0    & 53.2    & 17.8    & 14.4    & 54.0    & 20.3    \\ 
\llama{} (7B)  & 15.8    & 54.6    & 24.0   & 16.8    & 57.5    & 20.1   & 14.0    & 55.4    & 23.3   & 12.3    & 51.1    & 17.8    & 15.9    & 57.3    & 22.3   & 14.7    & 55.9    & 19.4    & 14.9    & 55.3    & 21.1    \\
\llama{} (13B) & 14.9    & 55.3    & 24.9   & 16.2    & 57.3    & 20.1   & 14.5    & 56.4    & 24.4   & 12.7    & 53.6    & 20.0    & 16.4    & 58.9    & 24.4   & 15.4    & 57.1    & 20.9    & 15.0    & 56.4    & 22.5   \\ \hline
\flan{} (3B)       & 15.0    & 57.7    & 11.1   & 9.3     & 52.5    & 6.1    & 10.5    & 56.0    & 7.5    & 10.8    & 54.9    & 7.6     & 9.8     & 55.7    & 6.2    & 9.3     & 53.2    & 6.0     & 10.8    & 55.0    & 7.4     \\
\mpt{} (7B) Instruct    & 23.2    & 64.5    & 22.4   & 14.5    & 58.1    & 13.4   & 15.0    & 61.1    & 15.9   & 14.0    & 56.8    & 12.9    & 14.8    & 60.5    & 16.1   & 12.9    & 57.1    & 13.1    & 15.8    & 59.7    & 15.6    \\
\falcon{} (7B) Instruct & 27.2    & 68.9    & 28.1   & 16.1    & 61.4    & 14.7   & 15.5    & 62.5    & 17.1   & 14.7    & 58.4    & 15.2    & 15.4    & 62.4    & 15.4   & 14.3    & 60.8    & 14.2    & 17.2    & 62.4    & 17.4    \\
\llama{} (7B) Chat      & 15.9    & 58.8    & 26.5   & 15.4    & 58.8    & 20.9   & 14.2    & 57.4    & 24.4   & 12.8    & 54.6    & 20.6    & 16.7    & 59.5    & 25.4   & 15.4    & 58.7    & 22.1    & 15.0    & 58.0    & 23.3    \\
\flan{} (11B)      & 16.3    & 58.8    & 12.2   & 10.8    & 55.5    & 7.5    & 10.8    & 57.3    & 8.2    & 12.3    & 56.1    & 9.1     & 9.7     & 55.2    & 6.3    & 9.0     & 54.9    & 5.9     & 11.5    & 56.3    & 8.2     \\
\llama{} (13B) Chat     & 16.2    & 59.2    & 27.5   & 15.8    & 59.0    & 21.4   & 14.2    & 57.2    & 24.3   & 13.0    & 54.7    & 21.2    & 16.7    & 58.9    & 24.8   & 15.5    & 58.7    & 22.4    & 15.3    & 58.0    & 23.6   \\ \hline
\flan{} (3B) (FT-QA)  & 26.6    & 66.2    & 25.2   & 16.1    & 55.0    & 16.9   & 15.4    & 58.2    & 16.4   & 11.7    & 53.8    & 10.5    & 12.6    & 55.7    & 12.0   & 12.8    & 52.2    & 12.7    & 15.9    & 56.8    & 15.6    \\
\falcon{} (7B) (FT-QA)     & 27.8    & 68.4    & 26.6   & 20.1    & 60.6    & 21.1   & 16.7    & 61.3    & 17.8   & 12.4    & 56.5    & 9.4     & 12.8    & 57.9    & 11.6   & 14.8    & 57.5    & 16.2    & 17.4    & 60.4    & 17.1    \\
\llama{} (7B) (FT-QA)    & 30.0    & 69.7    & 28.2   & 18.3    & 60.7    & 19.2   & 16.9    & 61.9    & 17.5   & 12.2    & 55.8    & 9.0     & 13.0    & 58.5    & 11.2   & 15.7    & 58.5    & 16.6    & 17.7    & 60.8    & 16.9    \\
\mpt{} (7B) (FT-QA)        & 28.9    & 69.0    & 27.6   & 18.6    & 59.6    & 20.6   & 16.4    & 61.0    & 17.5   & 12.9    & 56.1    & 10.7    & 13.1    & 57.6    & 11.5   & 14.0    & 56.5    & 15.4    & 17.3    & 60.0    & 17.2    \\ \hline
\flan{} (3B) (FT-All) & 27.8    & 67.4    & 25.7   & 16.0    & 55.8    & 17.1   & 15.5    & 59.3    & 15.3   & 11.4    & 54.5    & 9.3     & 11.7    & 55.7    & 10.4   & 13.0    & 53.1    & 13.1    & 15.9    & 57.6    & 15.2    \\
\falcon{} (7B) (FT-All)    & 27.3    & 68.6    & 26.1   & 18.9    & 59.9    & 19.8   & 16.1    & 61.0    & 16.7   & 11.7    & 55.4    & 8.0     & 12.8    & 58.0    & 10.9   & 14.8    & 57.5    & 16.5    & 16.9    & 60.1    & 16.3    \\
\llama{} (7B) (FT-All)   & 30.2    & 69.7    & 27.8   & 17.9    & 60.4    & 17.9   & 17.3    & 61.9    & 17.7   & 12.4    & 54.9    & 9.9     & 13.3    & 58.3    & 12.2   & 15.0    & 57.7    & 15.5    & 17.7    & 60.5    & 16.8 \\ 
\mpt{} (7B) (FT-All)       & 29.1    & 68.8    & 27.4   & 18.2    & 59.2    & 20.4   & 16.5    & 61.5    & 17.0   & 13.4    & 56.4    & 11.5    & 13.5    & 57.5    & 12.3   & 14.5    & 56.7    & 16.6    & 17.5    & 60.0    & 17.5    \\ \hline
ChatDoctor    & 26.2    & 68.2    & 28.8   & 15.8    & 61.3    & 16.0   & 16.1    & 62.6    & 18.6   & 15.2    & 58.9    & 15.6    & 16.5    & 62.9    & 18.2   & 14.8    & 60.2    & 15.0    & 17.4    & 62.3    & 18.7    \\
MedAlpaca 7B  & 26.4    & 67.8    & 27.1   & 14.7    & 55.6    & 13.0   & 13.4    & 59.3    & 15.0   & 12.3    & 55.1    & 12.6    & 13.9    & 59.0    & 15.4   & 12.5    & 56.8    & 10.2    & 15.5    & 58.9    & 15.6    \\
PMC LLama 13B & 19.7    & 62.6    & 20.9   & 12.7    & 55.8    & 11.0   & 13.5    & 58.8    & 14.4   & 45.6    & 70.7    & 43.6    & 14.8    & 59.6    & 14.0   & 11.9    & 57.0    & 10.1    & 19.7    & 60.7    & 19.0   \\ \hline

\end{tabular}
}
\caption{\aqa{} scores of base, instruction-tuned \llm{s} in the zero-shot setting,  \llm{s} fine-tuned with QLora and other biomedical and clinical instruction tuned models such as \chatdoctor{} (7B), \medalpaca{} (7B), \pmcllama{} (13B). FT-QA refers to models fine-tuned only with \aqa{} data and FT-All refers to models fine-tuned with both \mcqa{} and \aqa{} data.}
\label{tab:aqa_scores}
\end{table*}

%% file: section/prompts/chatdoctor_prompts.tex
\begin{figure*}[ht!]
\centering    
\begin{tcolorbox}[title=\mcqa{} Prompt for \chatdoctor{}]
Below is an instruction that describes a task, paired with an input that provides further context. Write a response that appropriately completes the request.\newline

\#\#\# Instruction:

If you are a doctor, please answer the medical questions based on the patient's description. Answer with the best option directly.\newline

\#\#\# Input:

Question: \{Example Question 1\}

Options:

A. \{Option Text\}

B. \{Option Text\}

C. \{Option Text\}

D. \{Option Text\}\newline

\#\#\# Response:

Answer:\{Correct Option\} \newline
...\newline
\#\#\# Instruction:

If you are a doctor, please answer the medical questions based on the patient's description. Answer with the best option directly.\newline

\#\#\# Input:

Question: \{Example Question K\}

Options:

A. \{Option Text\}

B. \{Option Text\}

C. \{Option Text\}

D. \{Option Text\}\newline

\#\#\# Response:

Answer:\{Correct Option\}\newline

\#\#\# Instruction:

If you are a doctor, please answer the medical questions based on the patient's description. Answer with the best option directly.\newline

\#\#\# Input:

Question: \{Question\}

Options:

A. \{Option Text\}

B. \{Option Text\}

C. \{Option Text\}

D. \{Option Text\}\newline

\#\#\# Response:

Answer:
\end{tcolorbox}
\caption{\mcqa{} prompt utilized for evaluating \chatdoctor{}. 5-shot prompting is utilized for the MMLU datasets whereas 1-shot prompting is utilized for all other \mcqa{} datasets.}
\end{figure*}

\begin{figure*}[ht!]
\centering    
\begin{tcolorbox}[title=Single Context \mcqa{} Prompt for \chatdoctor{}]
Below is an instruction that describes a task, paired with an input that provides further context. Write a response that appropriately completes the request.\newline

\#\#\# Instruction:

If you are a doctor, please answer the medical questions based on the patient's description. Analyze the question given its context. Answer with the best option directly.\newline

\#\#\# Input:

Context: \{Context Paragraph\}

Question: \{Example Question\}

Options:

A. \{Option Text\}

B. \{Option Text\}\newline

\#\#\# Response:

Answer:\{Correct Option\}\newline

\#\#\# Instruction:

If you are a doctor, please answer the medical questions based on the patient's description. Analyze the question given its context. Answer with the best option directly.\newline

\#\#\# Input:

Context: \{Context Paragraph\}

Question: \{Question\}

Options:

A. \{Option Text\}

B. \{Option Text\}\newline

\#\#\# Response:

Answer:
\end{tcolorbox}
\caption{Single Context \mcqa{} prompt utilized for evaluating \chatdoctor{} on the \processbank{} dataset.}
\end{figure*}

\begin{figure*}[ht!]
\centering    
\begin{tcolorbox}[title=Multi Context \mcqa{} Prompt for \chatdoctor{}]
Below is an instruction that describes a task, paired with an input that provides further context. Write a response that appropriately completes the request.\newline

\#\#\# Instruction:

If you are a doctor, please answer the medical questions based on the patient's description. Analyze the question given its context. Answer with the best option directly.\newline

\#\#\# Input:

Contexts: \{Context Paragraph 1\}

\{Context Paragraph 2\}

...

\{Context Paragraph N\}

Question: \{Example Question\}

Options:

A. \{Option Text\}

B. \{Option Text\}

C. \{Option Text\}\newline

\#\#\# Response:

Answer:\{Correct Option\}\newline

\#\#\# Instruction:

If you are a doctor, please answer the medical questions based on the patient's description. Analyze the question given its context. Answer with the best option directly.\newline

\#\#\# Input:

Contexts: \{Context Paragraph 1\}

\{Context Paragraph 2\}

...

\{Context Paragraph N\}

Question: \{Question\}

Options:

A. \{Option Text\}

B. \{Option Text\}

C. \{Option Text\}\newline

\#\#\# Response:

Answer:
\end{tcolorbox}
\caption{Multi-Context \mcqa{} prompt utilized for evaluating \chatdoctor{} on the \pubmedqa{} dataset.}
\end{figure*}

\begin{figure*}[ht!]
\centering    
\begin{tcolorbox}[title=Cloze \mcqa{} Prompt for \chatdoctor{}]
Below is an instruction that describes a task, paired with an input that provides further context. Write a response that appropriately completes the request.\newline

\#\#\# Instruction:

If you are a doctor, please answer the medical questions based on the patient's description. Analyze the question given its context. Pick the right option that corresponds to the XXXX in the question\newline

\#\#\# Input:

Context: \{Context Paragraph\}

Question: \{Question\}

Options:

A. \{Option Text\}

B. \{Option Text\}\newline

\#\#\# Response:

Answer:\{Correct Option\}\newline

\#\#\# Instruction:

If you are a doctor, please answer the medical questions based on the patient's description. Analyze the question given its context. Pick the right option that corresponds to the XXXX in the question\newline

\#\#\# Input:

Context: \{Context Paragraph\}

Question: \{Question\}

Options:

A. \{Option Text\}

B. \{Option Text\}\newline

\#\#\# Response:

Answer:
\end{tcolorbox}
\caption{Cloze \mcqa{} prompt utilized for evaluating \chatdoctor{} on the \biomrc{} datasets in settings A and B.}
\end{figure*}

\begin{figure*}[ht!]
\centering    
\begin{tcolorbox}[title=\aqa{} Prompt for \chatdoctor{}]
Below is an instruction that describes a task, paired with an input that provides further context. Write a response that appropriately completes the request.\newline

\#\#\# Instruction:

If you are a doctor, please answer the medical questions based on the patient's description.\newline

\#\#\# Input:

\{Question\}\newline

\#\#\# Response:\newline

\end{tcolorbox}
\caption{\aqa{} prompt utilized for evaluating \chatdoctor{}.}
\end{figure*}

%% file: section/prompts/medalpaca_prompts.tex
\begin{figure*}[ht!]
\centering    
\begin{tcolorbox}[title=\mcqa{} Prompt for \medalpaca{}]
Below is an instruction that describes a task, paired with an input that provides further context. Write a response that appropriately completes the request.
\newline

\#\#\# Instruction:

Answer this multiple-choice question.\newline

\#\#\# Input:

\{Question\}

A: \{Option Text\}

B: \{Option Text\}

C: \{Option Text\}

D: \{Option Text\}
\newline

\#\#\# Response:

The Answer to the question is:
\end{tcolorbox}
\caption{\mcqa{} prompt utilized for evaluating \medalpaca{}.}
\end{figure*}

\begin{figure*}[ht!]
\centering    
\begin{tcolorbox}[title=Single Context \mcqa{} Prompt for \medalpaca{}]
Below is an instruction that describes a task, paired with an input that provides further context. Write a response that appropriately completes the request.\newline

\#\#\# Instruction:

Analyze the question given its context. Answer this multiple-choice question.\newline

\#\#\# Input:

Context: \{Context Paragraph\}\newline

\{Question\}

A: \{Option Text\}

B: \{Option Text\}\newline

\#\#\# Response:

The Answer to the question is:
\end{tcolorbox}
\caption{Single Context \mcqa{} prompt utilized for evaluating \medalpaca{} on the \processbank{} dataset}
\end{figure*}

\begin{figure*}[ht!]
\centering    
\begin{tcolorbox}[title=Multi Context \mcqa{} Prompt for \medalpaca{}]
Below is an instruction that describes a task, paired with an input that provides further context. Write a response that appropriately completes the request.\newline

\#\#\# Instruction:

Analyze the question given its context. Answer this multiple-choice question.\newline

\#\#\# Input:

Contexts: \{Context Paragraph 1\}

\{Context Paragraph 2\}

...

\{Context Paragraph N\}\newline

\{Question\}

A: \{Option Text\}

B: \{Option Text\}

C: \{Option Text\}\newline

\#\#\# Response:

The Answer to the question is:
\end{tcolorbox}
\caption{Multi-Context \mcqa{} prompt utilized for evaluating \medalpaca{} on the \pubmedqa{} dataset.}
\end{figure*}

\begin{figure*}[ht!]
\centering    
\begin{tcolorbox}[title=Cloze \mcqa{} Prompt for \medalpaca{}]
Below is an instruction that describes a task, paired with an input that provides further context. Write a response that appropriately completes the request.\newline

\#\#\# Instruction:

Analyze the question given its context. Pick the right option that corresponds to the XXXX in the question.
\newline

\#\#\# Input:

Context: \{Context Paragraph\}\newline

\{Question\}

A: \{Option Text\}

B: \{Option Text\}

C: \{Option Text\}

D: \{Option Text\}\newline

\#\#\# Response:

The Answer to the question is:
\end{tcolorbox}
\caption{Cloze \mcqa{} prompt utilized for evaluating \medalpaca{} on the \biomrc{} datasets in settings A and B.}
\end{figure*}

\begin{figure*}[ht!]
\centering    
\begin{tcolorbox}[title=\aqa{} Prompt for \medalpaca{}]
Below is an instruction that describes a task, paired with an input that provides further context. Write a response that appropriately completes the request.\newline

\#\#\# Instruction:

Answer this question truthfully\newline

\#\#\# Input:

\{Question\}\newline

\#\#\# Response:\newline

\end{tcolorbox}
\caption{\aqa{} prompt utilized for evaluating \medalpaca{}.}
\end{figure*}

%% file: section/prompts/pmc_llama_prompts.tex
\begin{figure*}[ht!]
\centering    
\begin{tcolorbox}[title=\mcqa{} Prompt for \pmcllama{}]
Below is an instruction that describes a task, paired with an input that provides further context. Write a response that appropriately completes the request.
\newline

\#\#\# Instruction:

You're a doctor, kindly address the medical queries according to the patient's account. Answer with the best option directly.
\newline

\#\#\# Input:

\#\#\#Question: \{Question\}

\#\#\#Options:

A. \{Option Text\}

B. \{Option Text\}

C. \{Option Text\}

D. \{Option Text\}
\newline

\#\#\# Response:

\#\#\#Answer:
\end{tcolorbox}
\caption{\mcqa{} prompt utilized for evaluating \pmcllama{}.}
\end{figure*}

\begin{figure*}[ht!]
\centering    
\begin{tcolorbox}[title=Single Context \mcqa{} Prompt for \pmcllama{}]
Below is an instruction that describes a task, paired with an input that provides further context. Write a response that appropriately completes the request.
\newline

\#\#\# Instruction:

You're a doctor, kindly address the medical queries according to the patient's account. Analyze the question given its context. Answer with the best option directly.\newline

\#\#\# Input:

\#\#\#Question: \{Question\}

\#\#\#Context: \{Context Paragraph\}

\#\#\#Options:

A. \{Option Text\}

B. \{Option Text\}
\newline

\#\#\# Response:

\#\#\#Answer:
\end{tcolorbox}
\caption{Single Context \mcqa{} prompt utilized for evaluating \pmcllama{} on the \processbank{} dataset.}
\end{figure*}

\begin{figure*}[ht!]
\centering    
\begin{tcolorbox}[title=Multi-Context \mcqa{} Prompt for \pmcllama{}]
Below is an instruction that describes a task, paired with an input that provides further context. Write a response that appropriately completes the request.\newline

\#\#\# Instruction:

You're a doctor, kindly address the medical queries according to the patient's account. Analyze the question given its context. Answer with the best option directly.\newline

\#\#\# Input:

\#\#\#Question: \{Question\}

\#\#\#Contexts: \{Context Paragraph 1\}

\{Context Paragraph 2\}

...

\{Context Paragraph N\}

\#\#\#Options:

A. \{Option Text\}

B. \{Option Text\}

C. \{Option Text\}\newline

\#\#\# Response:

\#\#\#Answer:
\end{tcolorbox}
\caption{Multi-Context \mcqa{} prompt utilized for evaluating \pmcllama{} on the \pubmedqa{} dataset.}
\end{figure*}

\begin{figure*}[ht!]
\centering    
\begin{tcolorbox}[title=Cloze \mcqa{} Prompt for \pmcllama{}]
Below is an instruction that describes a task, paired with an input that provides further context. Write a response that appropriately completes the request.\newline

\#\#\# Instruction:

You're a doctor, kindly address the medical queries according to the patient's account. Analyze the question given its context. Pick the right option that corresponds to the XXXX in the question\newline

\#\#\# Input:

\#\#\#Question: \{Question\}

\#\#\#Context: \{Context Paragraph\}

\#\#\#Options:

A. \{Option Text\}

B. \{Option Text\}
\newline

\#\#\# Response:

\#\#\#Answer:
\end{tcolorbox}
\caption{Cloze \mcqa{} prompt utilized for evaluating \pmcllama{} on the \biomrc{} datasets in settings A and B.}
\end{figure*}

\begin{figure*}[ht!]
\centering    
\begin{tcolorbox}[title=\aqa{} Prompt for \pmcllama{}]
Below is an instruction that describes a task, paired with an input that provides further context. Write a response that appropriately completes the request.\newline

\#\#\# Instruction:

You're a doctor, kindly address the medical queries according to the patient's account.\newline

\#\#\# Input:

\#\#\#Question: \{Question\}\newline

\#\#\# Response:

\#\#\#Answer:
\end{tcolorbox}
\caption{\aqa{} prompt utilized for evaluating \pmcllama{}.}
\end{figure*}